%% file: main_arxiv.tex
\documentclass{article}

\usepackage{arxiv}

\usepackage{amsfonts}
\usepackage{amsmath,bm}
\usepackage{amssymb}
\usepackage{lineno}
\usepackage{tabularx}
\usepackage{multirow}
\usepackage{multicol}
\usepackage{subfig}
\usepackage{graphicx}
\usepackage{caption}
\usepackage{tikz}
\usepackage{color}
\usepackage{xspace}
\usepackage{enumerate}
\usepackage{xurl}
\usepackage[colorlinks,allcolors=blue]{hyperref}
\usepackage{xspace}
\usepackage[normalem]{ulem}

\usepackage[acronym, shortcuts]{glossaries}

\usepackage{colortbl}

\newcommand{\highest}[1]{{\textcolor{blue}{\mathbf{{#1}}}}}
\newcommand{\secondhighest}[1]{{\mathbf{{#1}}}}%

\newcolumntype{L}{>{\raggedright\arraybackslash}X}
\newcolumntype{C}{>{\centering\arraybackslash}X}
\newcolumntype{R}{>{\raggedleft\arraybackslash}X}

\newcommand{\mat}[1]{\mbox{\fontencoding{T1}\sffamily\slshape{#1\/}}} 
\renewcommand{\vec}[1]{\mbox{\textbf{#1}} }
\newcommand{\vecsymb}[1]{\boldsymbol{#1}}

\newcommand{\func}[1]{{\mbox{\usefont{OT1}{pzc}{m}{it}{#1}}}}
\newcommand{\set}[1]{\mathcal{#1}}

\newcommand\rurl[1]{%
  \href{https://#1}{\nolinkurl{#1}}%
}

\newacronym{lstm}{LSTM}{Long-Short Term Memory}

\title{In the Search for Optimal Multi-view Learning Models for Crop Classification with Global Remote Sensing Data}

\date{} 					

\author{ 
    Francisco Mena$^{1,2}$\href{https://orcid.org/0000-0002-5004-6571}{\includegraphics[scale=0.06]{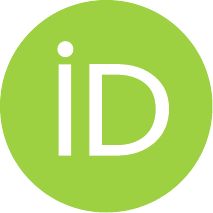}\hspace{1mm}},
	Diego Arenas$^{2}$\href{https://orcid.org/0000-0001-7829-6102}{\includegraphics[scale=0.06]{imgs/orcid.pdf}\hspace{1mm}}, 
	Andreas Dengel$^{1,2}$\href{https://orcid.org/0000-0002-6100-8255}{\includegraphics[scale=0.06]{imgs/orcid.pdf}\hspace{1mm}} \\
	$^1$Department of Computer Science,	University of Kaiserslautern-Landau (RPTU), Kaiserslautern, Germany;\\
	$^2$SDS, German Research Center for Artificial Intelligence (DFKI), Kaiserslautern, Germany.\\
	\texttt{f.menat@rptu.de}\\ 
}

\renewcommand{\headeright}{A Preprint}
\renewcommand{\undertitle}{Preprint submitted to Journal}
\renewcommand{\shorttitle}{Optimal Multi-view Learning for Crop Classification}

\hypersetup{
pdftitle={Optimal Multi-view Learning for Crop Clasification},
pdfauthor={Francisco Mena, Diego Arenas, Andreas Dengel},
pdfkeywords={Multi-view Learning, Multi-modal Learning, Data Fusion, Remote Sensing, Deep Learning, Supervised Learning}
}

\newcommand*\citep[1]{\cite{#1}}

\begin{document}

\maketitle

\input{content/abstract}


\input{content/introduction}
\input{content/relatedwork}
\input{content/methods}
\input{content/data}

\input{content/experiments}
\input{content/discussion}

\input{content/conclusion}

\input{content/end_arxiv}


\bibliographystyle{apalike}
\bibliography{main_arxiv}

\clearpage
\setcounter{table}{0}
\renewcommand{\thetable}{A\arabic{table}}
\setcounter{figure}{0}
\renewcommand{\thefigure}{A\arabic{figure}}
\appendix
\input{content/supp}

\end{document}

%% file: content/abstract.tex
\begin{abstract}

Studying and analyzing cropland is a difficult task due to its dynamic and heterogeneous growth behavior.
Usually, diverse data sources can be collected for its estimation. 
Although deep learning models have proven to excel in the crop classification task, they face substantial challenges when dealing with multiple inputs, named Multi-View Learning (MVL).
The methods used in the MVL scenario can be structured based on the encoder architecture, the fusion strategy, and the optimization technique.
The literature has primarily focused on using specific encoder architectures for local regions, lacking a deeper exploration of other components in the MVL methodology.
In contrast, we investigate the simultaneous selection of the fusion strategy and encoder architecture, assessing global-scale cropland and crop-type classifications. 
We use a range of five fusion strategies (Input, Feature, Decision, Ensemble, Hybrid) and five temporal encoders (LSTM, GRU, TempCNN, TAE, L-TAE) as possible configurations in the MVL method. 
We use the CropHarvest dataset for validation, which provides optical, radar, weather time series, and topographic information as input data.
We found that in scenarios with a limited number of labeled samples, a unique configuration is insufficient for all the cases. Instead, a specialized combination should be meticulously sought, including an encoder and fusion strategy.
To streamline this search process, we suggest identifying the optimal encoder architecture tailored for a particular fusion strategy, and then determining the most suitable fusion strategy for the classification task.
We provide a methodological framework for researchers exploring crop classification through an MVL methodology. 
\end{abstract}

%% file: content/introduction.tex
\section{Introduction} \label{sec:intro}

Accurate cropland maps are essential for assessing the climate effects on agriculture, food security, and resource management \cite{schneider2023eurocrops}. 
These croplands, representing farms or fields, often come as points or polygons. 
It has been shown that Remote Sensing (RS) data sources can be used as predictors for cropland classification or segmentation \cite{inglada2016improved,torbick2018fusion},
where deep learning (based on neural network models) is the predominant data-driven solution \cite{ofori-ampofo2021crop,saintefaregarnot2022multi,russwurm2020self}. 
The instance classification involves assigning a label to a particular geographical region, like a field, while the semantic segmentation assigns labels to multiple (small) geographical regions (referred to as pixels) within a larger region, like a farm.
These tasks can involve studying the whole crop growing season, from seeding to harvesting, leading to time series of varying lengths.

Learning from RS-based time series data of variable length presents unique modeling challenges. 
Particularly, when it comes to determining the optimal feature extraction technique regarding the accuracy \cite{russwurm2020self}.
An encoder can extract information from the entire time series. Variations of Neural Network (NN) architectures are usually used to learn from this type of data.  
For instance, Garnot et al. \cite{garnot2020satellite} show that useful representations of Satellite Image Time Series (SITS) for crop classification can be obtained with a tailored NN model that extracts spatial information at each time step, followed by a temporal aggregation.
Other works focus only on the temporal change and break the SITS into time series of pixels for pixel-wise mapping. 
For instance, standard encoder architectures in the literature make use of Recurrent Neural Network (RNN) with Long-Short Term Memory (LSTM, \cite{russwurm2017temporal}) or Gated Recurrent Unit (GRU, \cite{garnot2019time}) modules, standard Convolutional Neural Network (CNN) with 1-dimensional convolution over time (TempCNN, \cite{pelletier2019temporal}), or transformer-based models with attention mechanisms \cite{vaswani2017attention}, like Temporal Attention Encoder (TAE, \cite{garnot2020satellite}) and Light TAE (L-TAE, \cite{garnot2020lightweight}).
These papers focus on building specialized models for temporal data types.

Nowadays, the availability and diversity of RS sources have increased the interest in using models with multiple data sources \cite{camps2021deep}.
The reason for using multiple data sources is to corroborate and complement the information among individual observations. 
The Multi-View Learning (MVL) scenario emerges when searching for an optimal approach to combine the information from these various input data \cite{yan2021deep}. 
This is a challenging scenario considering the heterogeneous nature of RS and Earth observation data \cite{mena2023common}, with differences in spectral bands (bandwidth and number of channels), calibration, and spatial and temporal resolutions.
For instance, the temporal information from an optical view (passive observation) differs from a radar view (active observation). 
The optical view is affected by clouds, while the radar view may be affected by the surface roughness.

The MVL scenario with RS data has been explored in the literature. For instance, research works comparing fusion strategies \cite{cuelarosa2018dense,ofori-ampofo2021crop,saintefaregarnot2022multi}, and a yearly data fusion contest hosted by the IEEE GRSS\footnote{\rurl{www.grss-ieee.org/technical-committees/image-analysis-and-data-fusion/} (Accessed March 14th, 2024)}.
However, it is not yet clear what the advantages of different approaches in MVL are or how different MVL models can be compared. For this reason, we present a methodological framework to compare different models in the MVL scenario and assess them empirically.
Thus, we explore five fusion strategies and five encoder architectures for time series data in a pixel-wise crop classification task. 
We select common methods from the literature and validate them in the CropHarvest dataset \cite{tseng2021crop}. 
This dataset contains labeled data points that are (sparsely) distributed across the globe between 2016 and 2021 with five associated input views: multi-spectral optical SITS, radar SITS, weather time series, Normalized Difference Vegetation Index (NDVI) time series, and topographic information. 
In the context of crop classification, our main research question is \textit{what are the advantages of different MVL models under the same methodological framework?}
Moreover, we address the following questions: 
1) \textit{How does the selection of an encoder architecture and a fusion strategy affect the correctness of MVL model predictions?} 
2) \textit{How much are the predictions conditioned by the fusion strategy given an encoder architecture?} 
And 
3) \textit{How can multiple input views affect the model's confidence?} 

Previously, we compared different fusion strategies using a GRU-based encoder in the CropHarvest dataset \cite{mena2023crop}. In that work, we focus on finding the best model combination to outperform state-of-the-art models for a particular benchmark subset. The data subset consists of a binary crop classifier for Kenya and Togo.
The main finding is that to achieve the best results, a data-dependent placement of the fusion within the model is required \cite{mena2023crop}.
However, in this work, we provide in-depth insights into the following contributions:
\begin{enumerate}
    \item We compare five state-of-the-art encoder architectures for time series data (LSTM, GRU, TempCNN, TAE, L-TAE) in combination with five fusion strategies. 
    We include four fusion strategies presented in Mena et al. \cite{mena2023crop} (Input, Feature, Decision, and Ensemble), and a Hybrid strategy, which places multiple levels of fusion in the same model (we use a mix of Feature and Decision fusions).
    \item A comprehensive analysis including Brazil as a benchmark region to the Kenya, Togo and global datasets for evaluating the binary cropland classification. We also include a subset labeled with multiple classes, evaluating the multi-crop classification task.
    \item A methodological framework to study and compare MVL models, based on Mena et al. \cite{mena2023common}, as well as a detailed analysis of the used dataset and results.    
\end{enumerate}

The remainder of this paper is organized as follows: Section~\ref{sec:related} presents the related work. While Section~\ref{sec:methods} introduces the methods to be compared, Section~\ref{sec:data} describes the datasets used to assess the methods.
The experiment results are shown in Section~\ref{sec:exp}, with a discussion in Section~\ref{sec:discussion}. Finally, the conclusion is in Section~\ref{sec:conclusion}.

%% file: content/relatedwork.tex
\section{Related Work} \label{sec:related}

In crop-related studies, it has been explored how to handle SITS of variable length with Single-View Learning (SVL) models. Single-view in the context of using a single RS data source.
For instance, Zhong et al. \cite{zhong2019deep} compare NNs for time series (1D CNN and LSTM) to classical machine learning models (gradient boosted decision tree, Random Forest (RF), and support vector machine). Here, Landsat 7 and 8 SITS are used for pixel-wise crop classification in a California county, USA. 
Garnot et al. \cite{garnot2019time} also compare different NN architectures for Sentinel-2 (S2) SITS in crop classification for Southern France. They use a ConvLSTM (an LSTM recurrent network with convolution as operators), and CNN+GRU (a 2D CNN applied to the image in each time-step followed by a GRU recurrent network).
Moreover, Russwurm et al. \cite{russwurm2020self} include transformer models to compare 1D CNN and LSTM for S2-based crop classification in the Bavarian state, Germany. Later, Yuan et al. \cite{yuan2022sits} introduce a transformer model adapted to SITS (called SITS-Former) using S2 data and validating in some USA states.
Additionally, Zhao et al. \cite{zhao2021evaluation} modify NN models to handle missing time-steps in S2 SITS (caused by clouds), with a crop segmentation use-case in a Chinese city.
In all these region-specific studies, the common outcome is that a more complex NN architecture leads to models with better accuracy.

When considering the challenging MVL scenario, the crop-related studies have explored a single fusion strategy. 
The most common approach to fuse optical and radar SITS is input-level fusion (merging the information before feeding it to a machine-learning model). In some cases, combined with classical machine learning, like RF \cite{inglada2016improved,torbick2018fusion,dobrinic2021sentinel}, or with NN models for time series, like with temporal attention \cite{weilandt2023early,ebel2023uncrtaints}. 
Nevertheless, some studies have explored alternative fusion strategies.
For instance, Gadiraju et al. \cite{gadiraju2020multi} employ feature-level fusion (merge at the intermediate layers of NN models) for pixel-wise crop classification across the USA, using a 2D CNN encoder for a NAIP image and an LSTM encoder for a MODIS time series.
Rustowicz et al. \cite{mrustowicz2019semantic} use decision-level fusion (merge at the output layers of NN models) for crop segmentation in South Sudan and Ghana with U-shape CNN+LSTM models using Sentinel-1 (S1), S2, and Planet SITS.
Liu et al. \cite{liu2018deep} merge the predictions of 2D CNN models trained over different angles of aerial images, known as ensemble-based aggregation, for wetlands classification in a Florida ranch, USA.
In these studies, a specialized MVL model is designed for the data used. 

In other RS-based studies, there have been more comparisons of MVL models. 
For instance, using the ISPRS Semantic Labeling Challenge with optical images and elevation maps from Germany. 
In this benchmark, Audebert et al. \cite{audebert2018rgb} compare the integration of fusion across all encoder layers compared to just fuse on the last one,
while Zhang et al. \cite{zhang2020hybrid} later showed that a combination of these approaches (as a hybrid fusion) obtained the best results.
In RS image classification, Hong et al. \cite{hong2021more} compare fusion strategies at different levels in two benchmark datasets, finding that merging closer to the output layer is more robust to missing sensors.
Furthermore, Ferrari et al. \cite{ferrari2023fusing} showed that feature-level fusion was more robust to cloudy scenarios than input and decision level fusion for deforestation segmentation in Brazil, with S2 and S1 SITS.

To the best of our knowledge, there are a few crop-related studies that compare MVL models \cite{cuelarosa2018dense,ofori-ampofo2021crop,saintefaregarnot2022multi,follath2024multi}. However, they use a single encoder architecture for two RS views (optical and radar) and are limited to specific regions, e.g. Brazil \cite{cuelarosa2018dense}, France \cite{ofori-ampofo2021crop,saintefaregarnot2022multi} or Germany \cite{follath2024multi}. 
In contrast, we use a global dataset (multiple regions) with five input views while comparing both, the encoder architectures and the fusion strategies, in MVL models for crop classification. 

%% file: content/methods.tex
\section{Methods} \label{sec:methods}

In this section, we introduce a methodological framework to compare different MVL models. The framework is based on the encoder architectures used, the fusion strategy employed, additional components, and the optimization.
These decisions instantiate MVL models with different numbers of parameters, which we compare conceptually in this section.
We present five standard encoder architectures for time series data and five commonly used fusion strategies.

\subsection{Problem Formulation}
Consider the multi-view input data $\set{X}^{(i)} = \left\{ \mat{X}_v^{(i)} \right\}_{v \in \set{V}}$ for a sample or pixel $i$, with $\set{V}$ the set of available views, and the corresponding predictive target for classification $y^{(i)} \in [K]$ (with $[K] = \{ 1, 2, \ldots K\}$). The views with temporal features correspond to a multivariate time series (tensor size $T_v \times D_v$), while the views with static features are multivariate data (with dimensionality $D_v$).
Let $\func{E}_v$ be the encoder for the view $v$ that maps the input features to a single high-level vector representation $\vec{z}_v^{(i)} \in \mathbb{R}^d$, while $\func{P}_v$ be a prediction \textit{head}\footnote{The concept ``head'' is used as additional NN layers that are added on top of the encoder for the predictive task.} for the view $v$ that maps the learned representation to the predicted target probabilities $\hat{y}^{(i)} = \hat{p}(y \mid \set{X}^{(i)}) \in \mathbb{R}^{K}$. 
Additionally, we consider $\func{E}_F$ and $\func{P}_F$ as the encoder and prediction head for the fused information, and $\func{merge}(\cdot)$ as a merge function applied to any tensor size, e.g. concatenate or average.

\subsection{Encoder Architectures} \label{sec:methods:encoders}
We use state-of-the-art architectures to process RS-based time series data. Two RNN architectures: LSTM \cite{hochreiter1997long} and GRU \cite{cho2014properties}, both previously used in crop classification \cite{russwurm2017temporal,metzger2021crop}. One CNN architecture: TempCNN \cite{pelletier2019temporal} (previously used in crop classification \cite{najjar2024data}), and two architectures based on multi-head attention mechanism: TAE \cite{garnot2020satellite} and L-TAE \cite{garnot2020lightweight} (previously used in crop classification \cite{ofori-ampofo2021crop,saintefaregarnot2022multi}). 
For temporal views, the input is a time series of the form $\mat{X}^{(i)}  =  [ \vec{x}_1^{(i)}, \vec{x}_{2}^{(i)}, \ldots, \vec{x}_{T}^{(i)} ]$, where $\vec{x}_{t}^{(i)} \in \mathbb{R}^{D}$ is the $t$-th observation.  

\paragraph{Recurrent-based architecture}
These encoders extract high-level representations at each time step $t$ in the form of a hidden state $\vec{h}_t^{(i)} \in \mathbb{R}^d$. 
Let $\func{F}$ be a recurrent unit or layer, like LSTM or GRU, and $\vec{h}_0 = \vec{0}$. Then, the hidden state at $t$ is calculated based on the input at that time step and the hidden state from the previous time step: $\vec{h}_t^{(i)} = \func{F}( \vec{x}_t^{(i)}, \vec{h}_{t-1}^{(i)})$. These layers can be stacked and create a deep network. The deeper layers take the hidden state from the precedent layer as input, i.e. it replaces $\vec{x}_t^{(i)}$ with $\vec{h}_t^{(i)}$ from the previous layer.  
Finally, to extract a single vector representation, we use the hidden state from the last time step: $\vec{z}^{(i)} = \vec{h}_T^{(i)}$.
These recurrent units use multiple parameterisable gates to update the hidden states; however, the GRU uses fewer gates than the LSTM. 

\paragraph{Convolution-based architecture}
These models extract high-level representations based on the temporal local neighborhood (temporal window). Let $\func{F}$ be a convolution block (including padding, activation function, and pooling operators), then, the hidden features are calculated as $\mat{H}^{(i)} = \func{F}( \mat{X}^{(i)})$.
As with recurrent layers, these blocks can be stacked to obtain a deep network.
Finally, to extract a single vector representation from the tensor $\mat{H}^{(i)}$, we flatten it: $\vec{z}^{(i)} = \func{flatten}(\mat{H}^{(i)})$.
The TempCNN used in our work stacks 1-dimensional convolution blocks applied along the temporal dimension.

\paragraph{Attention-based architecture}
These encoders use the attention mechanism and positional encoding to create three vector representations at each time step $t$: the key ($\vec{k}_t^{(i)}$), query ($\vec{q}_t^{(i)}$), and value ($\vec{v}_t^{(i)}$). These representations are calculated with Multi-Layer Perceptrons (MLPs, i.e. fully connected layers) $\func{F}$, feeding with the input data $\vec{x}_t^{(i)}$ and corresponding positional encoding $\vec{p}_t$ at each time: $\vec{k}_t^{(i)}, \vec{q}_t^{(i)}, \vec{v}_t^{(i)} = \func{F}(\vec{x}_t^{(i)} + \vec{p}_t )$. 
Finally, to extract a single vector representation, a master query can be used as in \cite{garnot2020satellite}: $\vec{q}_M^{(i)} = \func{average}( \{ \vec{q}_t^{(i)} \}_{t=1}^{T} )$. Then, the attention mechanism works as a weighted average over the value vector, $\vec{z}^{(i)} = \sum_t \alpha_t^{(i)}  \cdot \vec{v}_t^{(i)}$, with $\vecsymb{\alpha}^{(i)} = \func{softmax}( \vec{q}_M^{(i)} \cdot \vec{k}_t^{(i)} / \sqrt{d} )$.
Multiple attention mechanisms are applied in parallel, i.e. multi-head, followed by another MLP.
The L-TAE reduces the number of parameters compared to TAE by encoding the master queries as a single learnable parameter: $ \vec{q}_M^{(i)} = \vec{q}_M \ \forall i$. 

\begin{figure*}[t]
\centering
\subfloat{\includegraphics[width=0.49\textwidth, page=1]{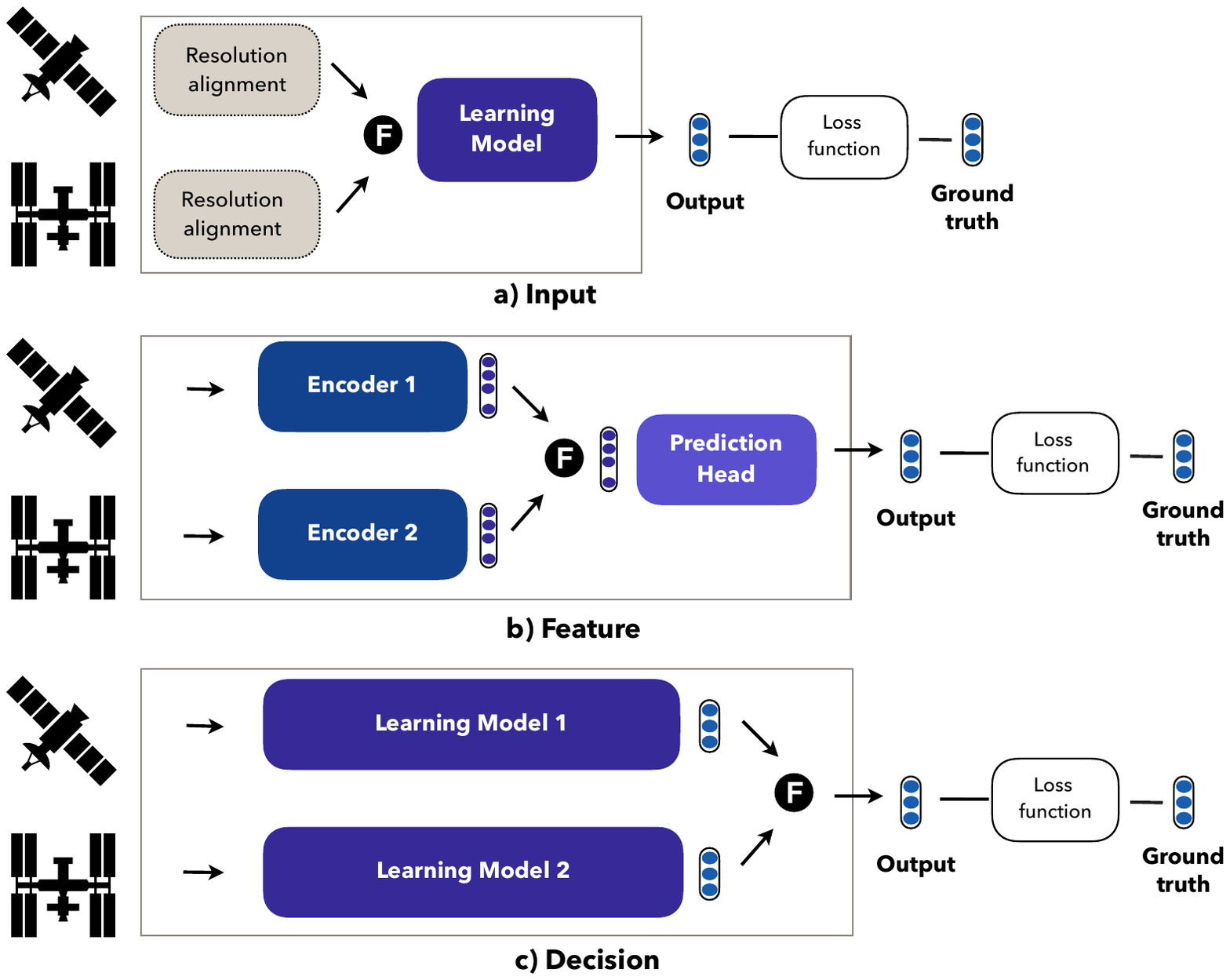}}
\hfill
\subfloat{\includegraphics[width=0.49\textwidth, page=3]{imgs/fusions_crop_class.pdf}}
\caption{Fusion strategies in MVL model compared in this manuscript. The first three (a, b, c) are the main fusion strategy distinctions made in the literature \cite{mena2023common}. The ``learning model'' is composed of an encoder followed by a prediction head.} \label{fig:methods}
\end{figure*}
\subsection{Fusion Strategies} \label{sec:methods:fusion}
We use common-choices of fusion strategies for MVL models \cite{ofori-ampofo2021crop,hong2021more,saintefaregarnot2022multi,mena2023common}: input-level fusion (Input in short), feature-level fusion (Feature in short), decision-level fusion (Decision in short), hybrid fusion (Hybrid in short), and ensemble-based aggregation (Ensemble in short). An illustration of these for two-view learning is presented in Fig.~\ref{fig:methods}.

\paragraph{Input fusion}
This strategy merges (concatenates) the views and feeds them to a single NN model (with encoder and prediction head) for prediction. As the input views can have different resolutions, an alignment step is used to match all the view dimensions, e.g. spatio-temporal alignment using re-sampling or interpolation operations. Considering the multi-view data for the $i$-th sample, this strategy is expressed by
\begin{align}
    \mat{X}_F^{(i)} &= \func{merge} \left( \func{alignment}( \left\{ \mat{X}_v^{(i)} \right\}_{v \in \set{V}} ) \right) , \\
    \hat{y}^{(i)} &= \func{P}_F \circ \func{E}_F ( \mat{X}_F^{(i)}).
\end{align}

\paragraph{Feature fusion}
This strategy uses a view-dedicated encoder to map each view to a new high-level feature space. Then, a merge function combines this information, obtaining a single joint representation. At last, a prediction head is used to generate the final prediction. Considering the multi-view data for the $i$-th sample, this is expressed by
\begin{align}
    \vec{z}_v^{(i)} &= \func{E}_v ( \mat{X}_v^{(i)} )  \ \ \forall v \in \set{V}  , \\
    \vec{z}_F^{(i)} &= \func{merge} \left( \left\{  \vec{z}_v^{(i)} \right\}_{v \in \set{V}} \right) , \\
    \hat{y}^{(i)} &= \func{P}_F ( \vec{z}_F^{(i)} ) .
\end{align}

\paragraph{Decision fusion}
This strategy utilizes a NN model (with an encoder and a prediction head) for each view. These view-dedicated models generate individual decisions, the crop probability. The outputs of these models are merged (usually averaging) to yield the aggregated prediction, similar to a mixture of experts \cite{masoudnia2014mixture}. 
Considering the data for the $i$-th sample, this is expressed by 
\begin{align}
    \label{eq:dec:pred} y_v^{(i)} &= \func{P}_v \circ \func{E}_v ( \mat{X}_v^{(i)})  , \ \ \forall v \in \set{V} ,  \\
    \label{eq:dec:agg} \hat{y}^{(i)} &= \func{merge} \left( \left\{  y_v^{(i)} \right\}_{v \in \set{V}}  \right) .
\end{align}

\paragraph{Hybrid fusion}
This fusion combines some above-mentioned strategies in the same MVL model. For instance, we consider the case of a feature-decision-level mix, which can also be considered two models with shared encoders per view. 
Considering the multi-view data for the $i$-th sample and average as the merge function, this is expressed by
\begin{align}
    \vec{z}_v^{(i)} &= \func{E}_{v} ( \mat{X}_v^{(i)}) \ \ \forall v \in \set{V} , \\ 
    y^{(i, \text{feature})} &= \func{P}^{\text{feature}}_F \circ \func{merge} \left( \left\{  \vec{z}_v^{(i)} \right\}_{v \in \set{V}} \right) , \\ 
    {y}^{(i, \text{decision})} &= \func{merge} \left( \left\{  \func{P}_{v}^{\text{decision}} ( \vec{z}_v^{(i)} ) \right\}_{v \in \set{V}}  \right) , \\
    \hat{y}^{(i)} &= \func{merge} \left( y^{(i,\text{feature})}, {y}^{(i,\text{decision})}   \right) .
\end{align}

Let $\func{L}(y^{(i)}, \hat{y}^{(i)})$ be a loss function, in our case the cross-entropy $\func{L}(y, \hat{y}) = - \sum_k y_k \log{\hat{y}_k} $, with $\hat{y}^{(i)}$ the prediction based on the fused multi-view data.
The previous strategies describe MVL models that are learned by minimizing the function of the form $\min_{ \left\{\func{P}_v, \func{E}_v\right\}_{v \in \set{V}} } \func{L}(y^{(i)}, \hat{y}^{(i)})$.

\paragraph{Ensemble aggregation}
Similar to Decision fusion, this strategy has a model yielding a decision per view. However, these view-dedicated models are trained independently without merging, i.e. $\min_{\func{P}_v, \func{E}_v} \func{L}(y^{(i)}, \hat{y}_v^{(i)}) \ \forall v \in \set{V}$, with $\hat{y}_v^{(i)}$ as in \eqref{eq:dec:pred}, i.e. the models can be completely different.
Then, at inference, the aggregated prediction is the average decision in this ensemble of trained models, as in \eqref{eq:dec:agg}.

\subsection{Comparison}
To summarize, Input fusion is the only strategy without view-dedicated models, and Ensemble aggregation is the only strategy that does not optimize over the fusion, i.e. it fuses the information without explicitly learning to fuse.
One could further analyze the conceptual differences of the fusion strategies, as detailed in Mena et al. \cite{mena2023common}.
However, we illustrate that there is an important distinction in the number of parameters obtained with these strategies and encoder architectures. 

In Table~\ref{tab:parameters} we show the number of parameters in each encoder architecture and view for the configurations used in our work. The architecture with the most parameters is the convolution-based one, with more than 13 times the number of parameters of the lightest architecture, L-TAE. 
For simplicity, let $n_{E}$ be the number of learnable parameters in the encoder $\func{E}$ (assuming all view-dedicated encoders have the same architecture), and $n_{P}$ the number of parameters in the prediction head $\func{P}$ (assuming same prediction heads). 
Then, using a pooling merge function, i.e. a merge function that keeps the same dimensionality of individual representations (e.g. average), the total number of parameters for each MVL model are: $n_{E} + n_{P}$ for Input, $|\set{V}| \cdot n_{E} + n_{P}$ for Feature, $|\set{V}| \cdot (n_{E} + n_{P})$ for Decision and Ensemble, and $|\set{V}|\cdot (n_{E} + n_{P}) + n_{P}$ for Hybrid strategies.

\subsection{Additional Components} \label{sec:methods:components}
In this work, we consider two components that can be included only with the Feature, Decision, and Hybrid strategies.
The first one is gated fusion (G-Fusion in short, \cite{mena2024adaptive}), a method that adaptively merges views ($\vec{z}_v^{(i)}$ or $y_v^{(i)}$) through a weighted sum: $\func{merge}( \{  \vec{z}_v^{(i)} \}_{v \in \set{V}} ) = \sum_{v \in \set{V}} \vecsymb{\alpha}_v^{(i)} \odot \vec{z}_v^{(i)}$. The weights $ \vecsymb{\alpha}_v^{(i)} $ have the same dimension as the learned representation (feature-specific weight) and are computed for each sample based on a gated unit, ${\alpha}^{(i)} = \func{GU}(\{  \mat{X}_v^{(i)} \}_{v \in \set{V}})$. 
We use the learned representations of the view-dedicated models as input for the gated unit \cite{mena2024adaptive}. 
The second is multiple losses (Multi-Loss in short, \cite{benedetti2018text}); this method includes one loss function for each view-specific prediction $\hat{y}_v^{(i)}$, to force the model to predict the target based on the individual information. 
These loss functions are added with a weight $\gamma$ to the final learning objective: $\func{L}(y^{(i)}, \hat{y}^{(i)}) + \gamma \sum_{v \in \set{V}} \func{L}(y^{(i)}, \hat{y}_v^{(i)})$. We use the value from the original proposal \cite{benedetti2018text}, i.e. $\gamma = 0.3$. 
For the Feature fusion, auxiliary prediction heads are included to generate the view-specific predictions: $\hat{y}_v^{(i)} = \func{P}_{v}^{\text{auxiliary}}(\vec{z}_v^{(i)})$.

%% file: content/data.tex
\section{Data} \label{sec:data}

\begin{figure}[!t]
    \centering
    \includegraphics[width=\linewidth]{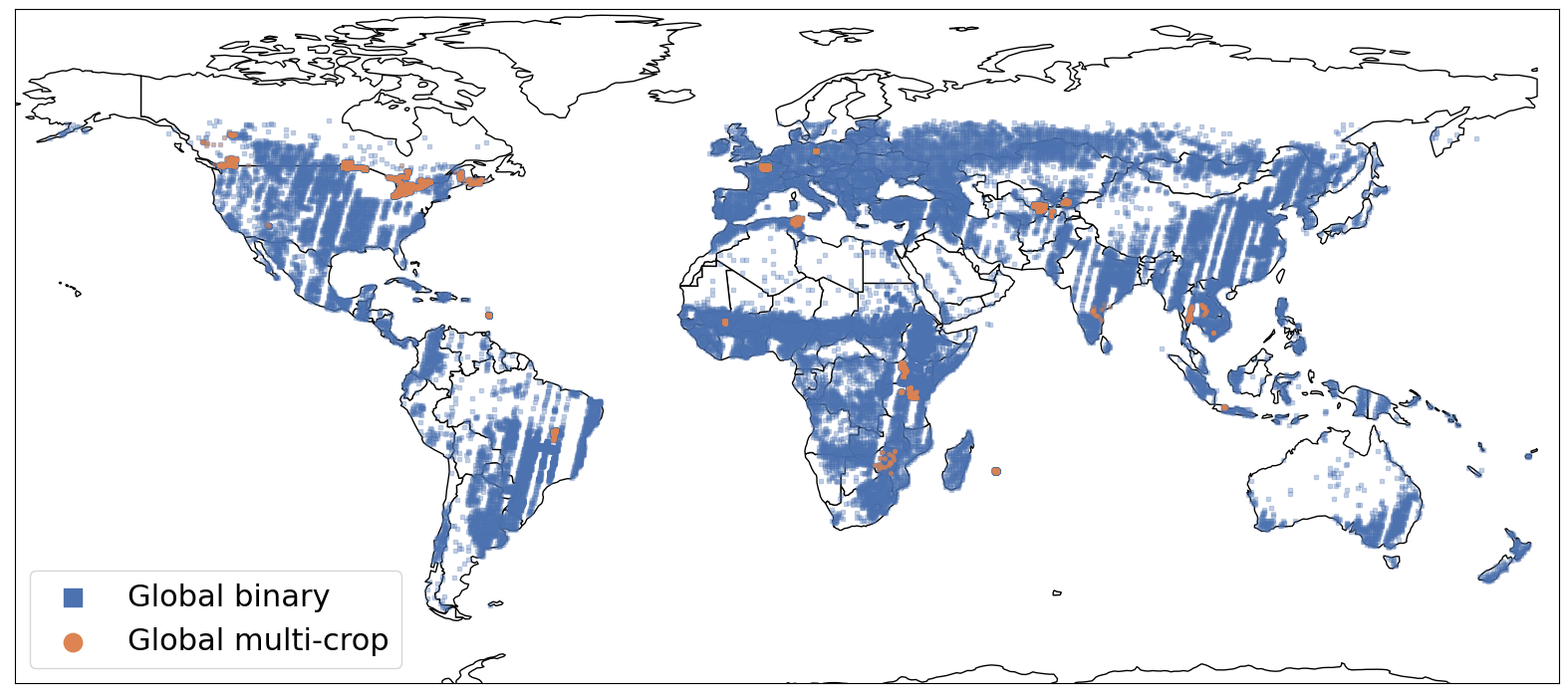}
    \caption{Spatial coverage of the CropHarvest dataset. The blue squares are the samples with binary labels (all global data), while the orange circles are the subset that has multiple crop-type labels (multi-crop subset).}\label{fig:data:globe}
\end{figure}
\begin{figure}[!t]
    \centering
    \subfloat[Data points per country in the Global Binary scenario. \label{fig:data:samplesdesc:country}]{\includegraphics[width=\linewidth]{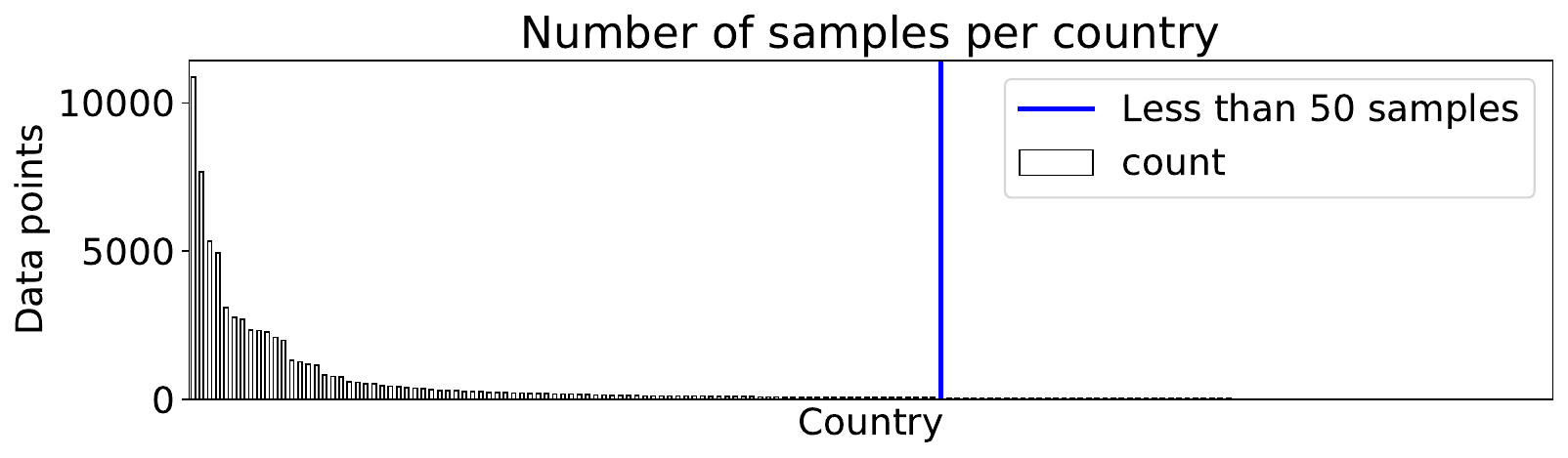}}\\
    \subfloat[Data points in the Global Multi-crop. \label{fig:data:samplesdesc:class}]{\includegraphics[width=0.6\linewidth]{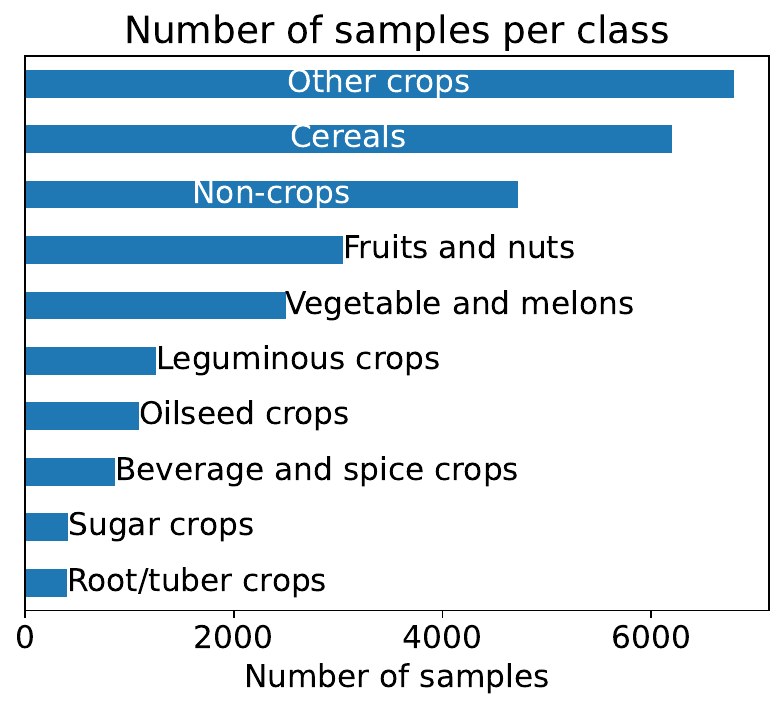}}
    \caption{Number of samples per country and crop-type in CropHarvest data.}\label{fig:data:samplesdesc}
\end{figure}

\begin{table}[!t]
    \centering
    \caption{Number of samples per year and continent in the CropHarvest data. NA/SA are abbreviations for North/South America.}\label{tab:data:year_cont}
    \small
    \begin{tabularx}{\linewidth}{l|RRRRRR} \hline
        {Year}  & {2016} & {2017} & {2018} & {2019} & {2020} & {2021} \\ 
        Samples &  27245 & 6418 & 5775 & 11388   &   6224 &   8195 \\ \hline \hline
        {Continent} & {Africa} & {Asia} & {Europe} & {NA} & {Oceania} & {SA} \\
        Samples & 16633 & 11039 & 19446 & 10656 &  756 & 6715\\
        \hline
    \end{tabularx}
\end{table}
\paragraph{Data description}
The case study corresponds to pixel-wise crop identification. Multiple RS sources are available as input data to detect if a specific crop (target crop) is growing in a particular coordinate (region). 
We use the CropHarvest dataset \cite{tseng2021crop} that has (sparsely) distributed data points across the Earth between 2016 and 2021. See Fig.~\ref{fig:data:globe} for the spatial coverage. 
These data points correspond to 65245 samples harmonized between points and polygons into a 100 $m^2$ region each (10 meters per pixel of spatial resolution), please refer to \cite{tseng2021crop} for this process. We filtered out samples that did not have associated RS data for prediction.
The number of samples per year and continent is shown in Table~\ref{tab:data:year_cont}.
The data per country is illustrated in Fig.~\ref{fig:data:samplesdesc:country}, where a long-tail distribution can be seen, with almost half of the countries with less than 50 samples. The five countries with the largest sample size are France, Canada, Brazil, Uzbekistan, and Germany, while the countries with fewer data are from Asia.
We selected this dataset for its global scale and large variability that reflect different points of view in the analysis.

\begin{table}[!t]
    \centering
    \caption{Number of samples in each data scenario. The percentage of positive data in the binary task is shown in parentheses. In Kenya, this corresponds to maize crops, in Brazil the coffee crops, while in Togo and Global Binary any crop.}\label{tab:data:samples}
    \small
    \begin{tabularx}{\linewidth}{L|RR} \hline
        {Data} & {Training  samples} & {Testing samples} \\ \hline
        Kenya &  1319 ($20.0\%$) & 898 ($64.0\%$) \\
        Brazil & 1583 ($1.00\%$) & 537454 ($32.4\%$) \\
        Togo  & 1290 ($55.0\%$) & 306 ($34.6\%$) \\
        \hline
        Global Binary & 45725 ($66.4\%$) & 19520 ($66.0\%$) \\ 
        Global Multi-crop &  19066 \hspace{1.5em} (---) & 8142 \hspace{1.5em} (---)\\
        \hline
    \end{tabularx}
\end{table}
\paragraph{Evaluation scenarios and target task}
The dataset provides three geographical regions as a benchmark for binary (cropland) classification, with the corresponding training and testing data. The tasks are to identify maize from other crops (maize vs the rest) in Kenya, coffee crops in Brazil, and distinguish between crop vs non-crop in Togo.
In these regions, the testing data comes from randomly selected polygons in space for a specific year.
Additionally, we create two scenarios that consider data from all the countries. The first scenario, Global Binary, uses all data for the cropland (crop vs. non-crop) classification, while the second, Global Multi-crop, uses the subset of samples that have a greater granularity in the labels for a multi-class (crop-type) classification.
The classes (9 groups of crop-types and one non-crop) and data distribution for the latter scenario are depicted in Fig.~\ref{fig:data:samplesdesc:class}.
The multi-crop is a subset of the binary set because not all samples have this finer label granularity.
For the global scenarios, we randomly select 30\% of the data (in space and time) for testing.
Finally, Table~\ref{tab:data:samples} displays the distribution of training and testing samples for each evaluation scenario\footnote{The testing data from the benchmark countries (Togo, Kenya and Brazil) is excluded in Figures.~\ref{fig:data:globe} and \ref{fig:data:samplesdesc}, and Table~\ref{tab:data:samples}.}. 

\begin{figure*}[t!]
    \centering
    \subfloat[Global data.]{\includegraphics[width=0.47\linewidth]{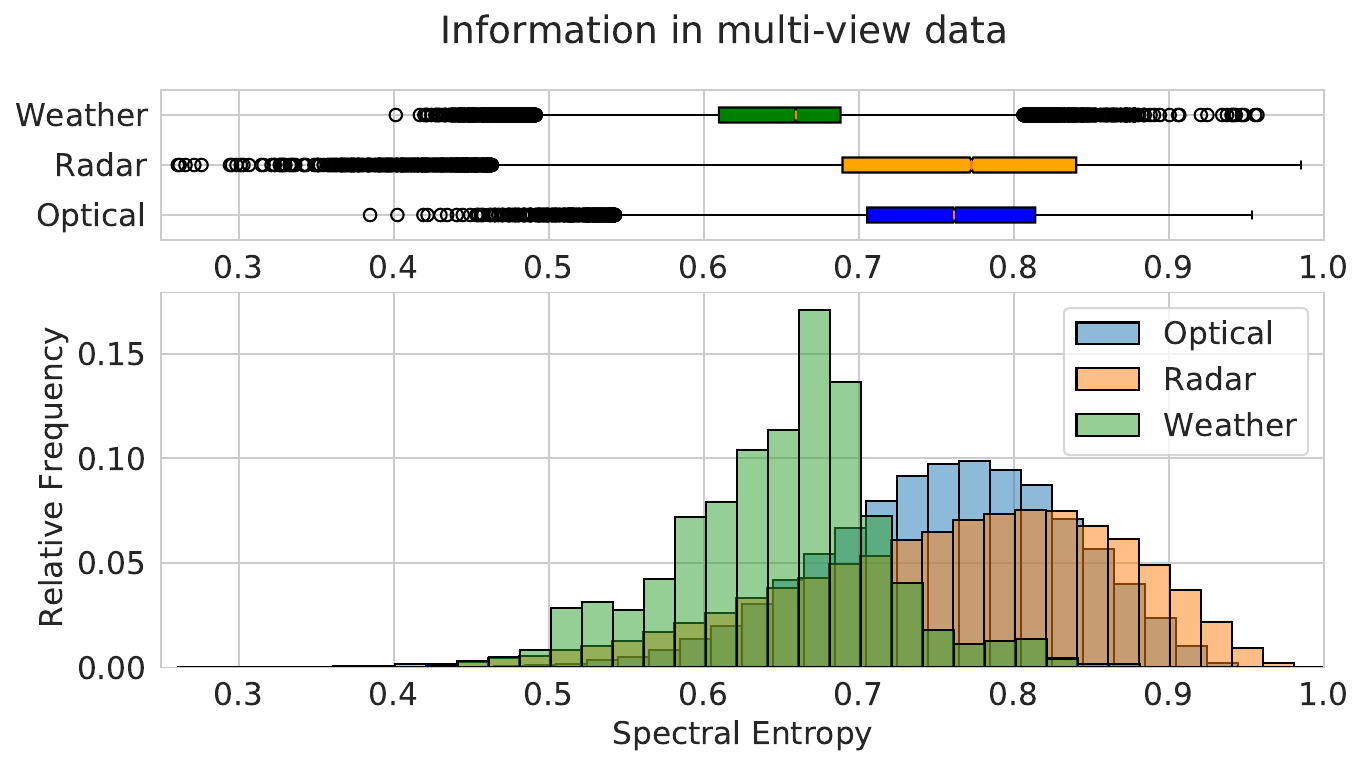}}\hfill\quad
    \subfloat[Kenya data.]{\includegraphics[width=0.47\linewidth]{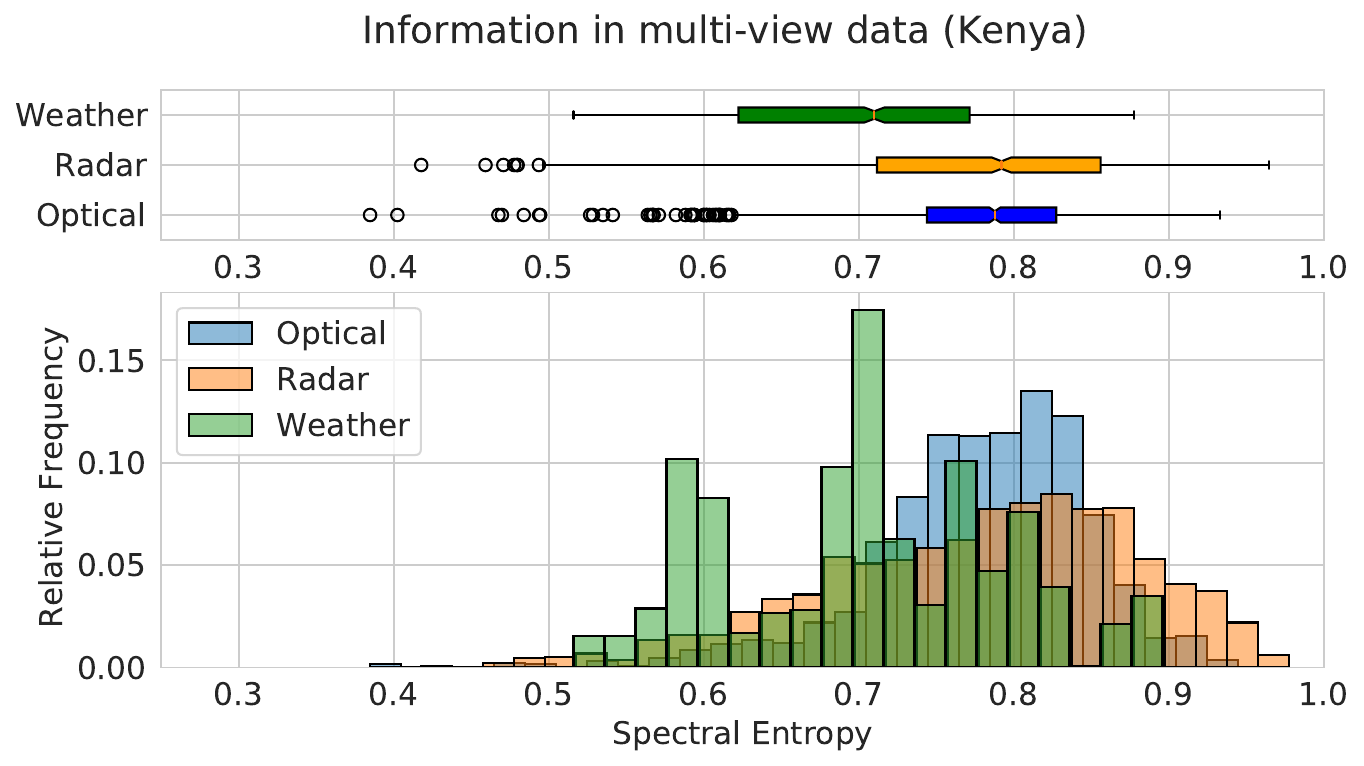}}\\
    \subfloat[Brazil data.]{\includegraphics[width=0.47\linewidth]{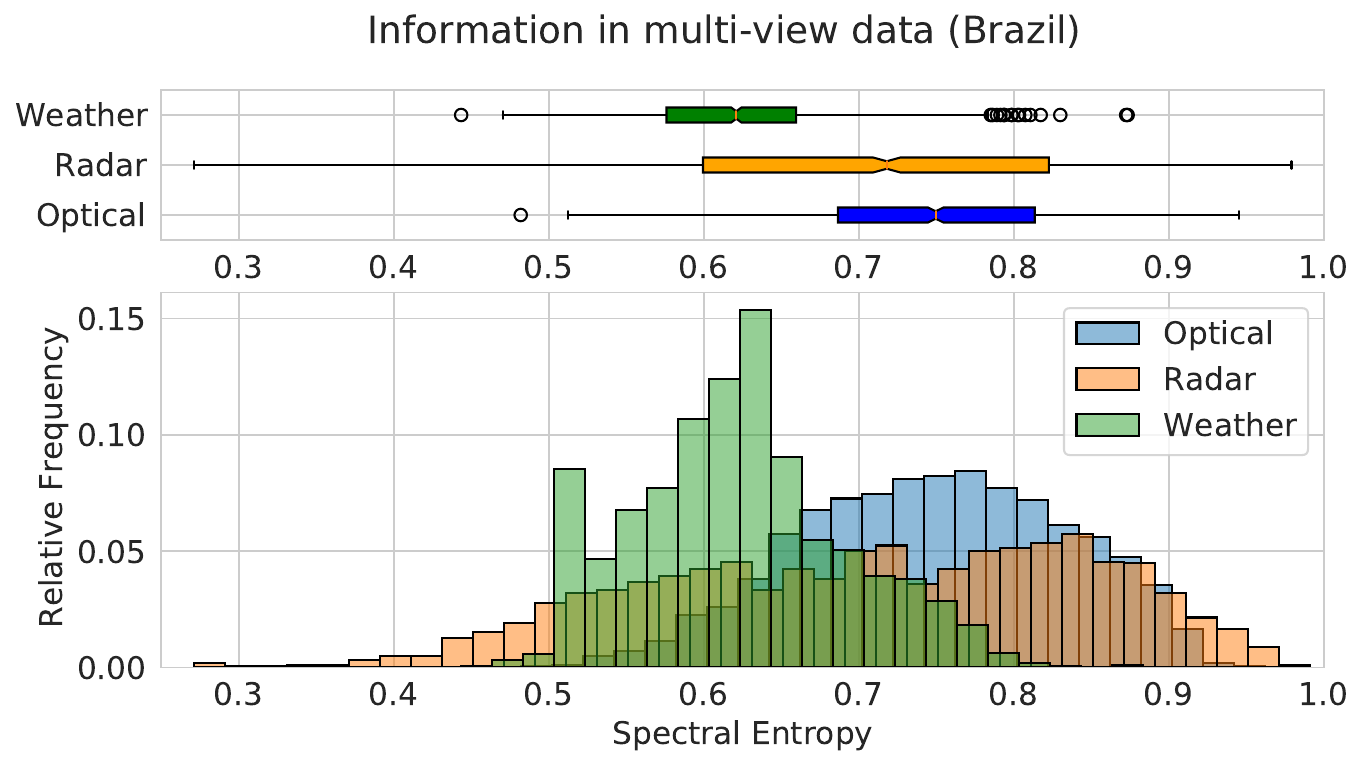}}\hfill\quad
    \subfloat[Togo data.]{\includegraphics[width=0.47\linewidth]{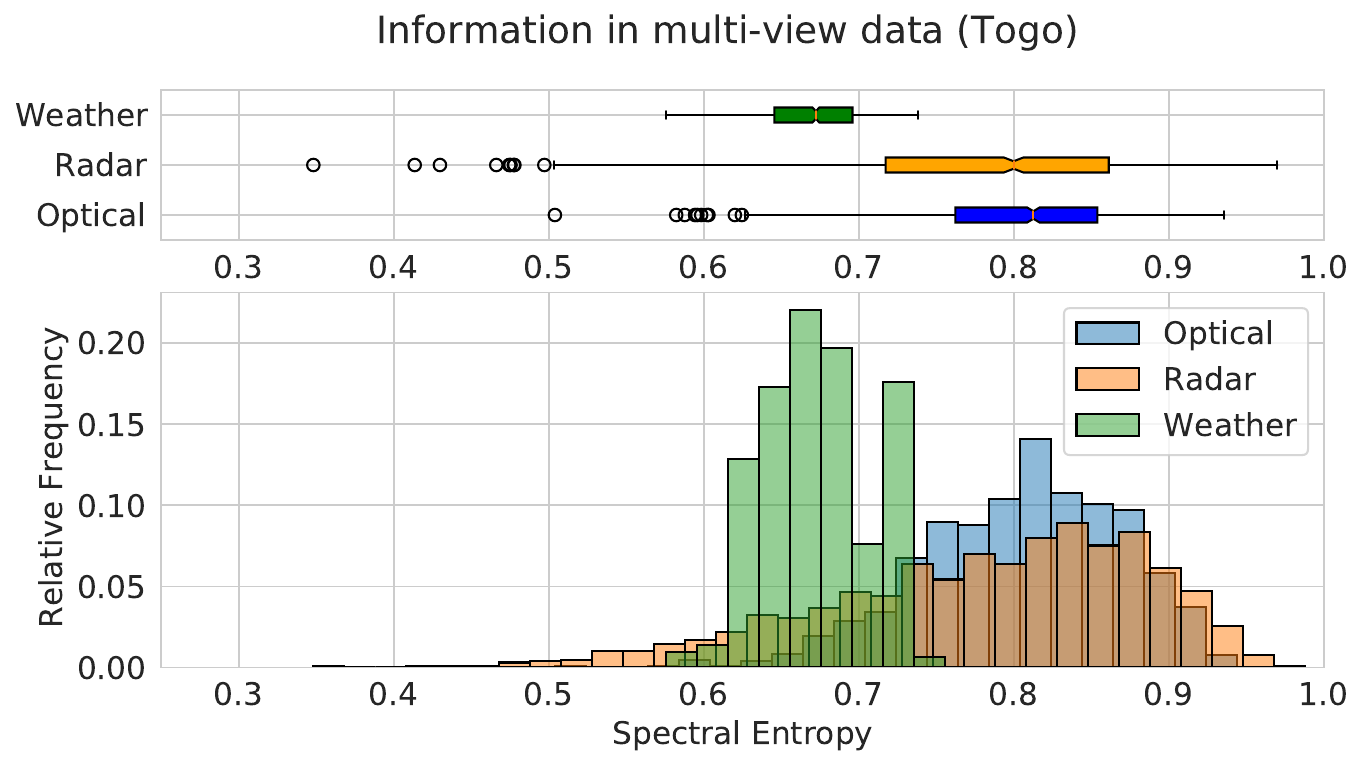}}
    \caption{Boxplot and histogram of the spectral entropy (proxy to non-periodic patterns) of the temporal views across the different scenarios.}
    \label{fig:entropy}
\end{figure*}
\paragraph{Multi-view input data}
Each labeled sample has five views as input data from four RS sources: S2, S1, ECMWF Reanalysis v5 (ERA5), and Shuttle Radar Topography Mission (SRTM).
The optical view is a multi-spectral optical SITS with 11 bands from the S2 mission at 10-60 m/px spatial resolution and approximately five days of revisiting time. 
The optical view 
provides the main information on the crop's composition, growth stage, canopy structure, and leaf water content \cite{tseng2021crop}.
The radar view is a 2-band (VV and VH polarization) radar SITS from the S1 mission at 10 m/px and a variable revisit time.
The radar view 
could penetrate cloud cover and provide information about the geometry and water content of the crop.
The weather view is a 2-band (precipitation and temperature) time series from the ERA5 at 31 km/px and hourly temporal resolution.
The weather view provides information about the expected crop development based on the climate conditions.
The NDVI view, which provides information about healthy and dense vegetation, is calculated from the optical view.
These temporal views are re-sampled monthly over one year.
The static information of elevation and slope is also included as a topographic view, coming from the SRTM's digital elevation model at 30 m/px. 
The topographic view can provide information about the suitability of certain crops \cite{tseng2021crop}.
The input views in the benchmark are spatially interpolated to a 10 m/px for a pixel-wise mapping, see \cite{tseng2021crop} for further details.

\paragraph{Descriptive analysis}
We visualize the time series information of the optical, radar, and weather views in Fig.~\ref{fig:entropy}. For this, the spectral entropy across the signal is calculated as a proxy for the information level \cite{inouye1991quantification}, i.e. a higher value is associated with a non-periodic pattern while a lower value means a constant-like or periodic-behavior signal. We calculated the mean of the spectral entropy across the features in each view.
This shows that \textit{a-priori}, the weather view is the one with less information, and that the radar view is the one with more stochastic and non-periodic patterns. 
In addition, the views on the Brazil data show a slightly lower mean entropy than in the other scenarios.

%% file: content/experiments.tex
\section{Experiments} \label{sec:exp}

\subsection{Evaluation Settings} \label{sec:exp:settings}

We assess the different MVL models (involving the encoder architectures and fusion strategies), repeating each experiment 20 times and reporting the metrics average with the standard deviation in the testing data. We use three metrics to assess the predictions in the classification task: average accuracy (AA), Kappa score ($\kappa$), and F1 macro score ($F_1^{\text{macro}}$):
\begin{align}
    & \text{AA} = \frac{1}{K} \sum_{k \in [K]} \frac{T_{P_k} + T_{N+k}}{T_{P_k} + F_{N_k} + T_{N_k} + F_{P_k}} ,   \\
    \kappa &= \frac{ 2 \cdot (T_P \cdot T_N - F_N \cdot F_P)}{ (T_P+F_P) (F_P+T_N) + (T_P+F_N) (F_N+T_N) } , \\
    & F_1^{\text{macro}}  = \frac{1}{K} \sum_{k \in [K]} 2 \frac{P_k \cdot R_k}{P_k + R_k} ,
\end{align}
with $T_{P_k}$ the false positive rate, $F_{N_k}$ the false negative rate, $T_{N_k}$ the true positive rate, $F_{P_k}$ the false positive, $P_k$ the precision, and $R_k$ the recall, all regarding class $k \in [K]$.

Throughout our experiments, we incorporated an SVL baseline composed of an NN (encoder and prediction head) trained using a single-view as input data. We report the results of the SVL model with the optical or radar view, selecting the one that yields the most favorable predictions in each case, while the results for both can be found in the appendix \ref{sec:appendix:tab}.

We experiment with the encoder architectures described in Sec.~\ref{sec:methods:encoders} for the temporal views (optical, radar, weather, and NDVI), while for the elevation view (static features) we only used an MLP as the encoder. We use the NDVI as a separate view, as has been shown optimal results in the literature \cite{audebert2018rgb,mrustowicz2019semantic,sheng2020effective}.
For the hyperparameter tuning in each encoder architecture, we did a random exploration of 10 to 50 trials in Kenya data. The hyperparameters tried for each architecture were the number of layers, the hidden state/features, the use of batch-normalization, in addition to more specific like bidirectional recurrence in GRU and LSTM, number of heads in TAE and L-TAE, and kernel size in TempCNN.
We use the configurations suggested in the original papers as the initial guesses of the parameters (in most cases, these are the best).
The reason for making the hyperparameter tuning using the Kenya data is that it was the focus of the Agriculture-vision competition\footnote{\rurl{www.agriculture-vision.com/agriculture-vision-2022/prize-challenge-2022} (Accessed March 14th, 2024)} and it is the more challenging evaluation scenario.
It is important to note that we selected the same encoder architecture for all the temporal views across the different model configurations, i.e. we did not test combinations such as LSTM for optical and GRU for radar.
Among the common selected hyperparameters in the encoders are: 64 dimensions in the hidden features with two layers, and 64 dimensions in the embedding vector. 
The prediction head is the same for all fusion strategies, an MLP with a single hidden layer of 64 units and a batch-normalization layer.
For regularization, we use 20\% of dropout. In Table \ref{tab:parameters} we compare the number of parameters of encoder architectures depending on the input view.

\begin{table}[!t]
    \centering
    \caption{Number of learnable parameters of NN architectures (encoders or prediction head) for different views.} \label{tab:parameters}
    \small
    \begin{tabularx}{\linewidth}{l|R|R|R|R|r} \cline{2-6}
         Encoder & Optical & Radar & Weather & NDVI & Topography  \\ \hline
         LSTM & 57152 & 54592 &  54592 &  50688 & \\
         GRU & 43904 & 42176 & 42176 & 41984 & \\
         TAE & 56598 & 56004 & 56004 & 55938 &  \\
         L-TAE & 19350 & 18756 & 18756 &  18690 &  \\ 
         TempCNN & 258880 & 256000 &  256000 & 255680 & \\  \cline{2-6}
         MLP &\multicolumn{4}{c|}{-} &  4352 \\ \hline
         \multicolumn{3}{c}{Prediction head} & \multicolumn{3}{c}{20802} \\ 
         \hline
    \end{tabularx}
\end{table}
We train the models from scratch with the ADAM optimizer \cite{kingma2014adam}, a learning rate of $10^{-3}$, and a batch size of 256. We use an early stopping criterion with 5 tolerance steps on a 10\% validation data extracted from the training data. 
We incorporate class weights, inversely proportional to class frequencies, into the objective function \cite{king2001logistic}. This approach addresses the class unbalance (Fig.~\ref{fig:data:samplesdesc:class}) by ensuring a balanced impact of the samples from each class within the loss function.

\subsection{Class Prediction Results} \label{sec:exp:results}

The first question we explore is \textit{how does the selection of an encoder architecture and a fusion strategy affect the correctness of MVL model predictions?} To address this, we run the five encoder architectures in the five fusion strategies described in Sec.~\ref{sec:methods:encoders} and ~\ref{sec:methods:fusion}, including the two components from Sec.~\ref{sec:methods:components} for the Feature, Decision, and Hybrid strategies. This generates 31 experiments ($5\cdot 5 + 2\cdot3$) that we repeated 20 times. 
The results for the country-specific evaluation are shown in Table \ref{tab:bestall:country}, and for the global in Table \ref{tab:bestall:global}.
Overall, we observe that the SVL classification results were outperformed by the MVL models in different amounts depending on the metric and evaluation scenario. For instance, the AA increased around $0.5$ points in Brazil, from $97.03$ to $97.50$, and $4$ points in the global evaluation, while the $\kappa$ score increased around $2$ points in Brazil and $8$ points in the Global Binary, from $55.05$ to $63.64$. 
As a common result in the literature, this suggests that additional RS sources help improve the classification. However, there is an exception in Kenya, where the LSTM encoder with the radar view obtains the best predictions. We further discuss this in Sec.~\ref{sec:discussion}.

\begin{table*}[!t]
    \centering
    \caption{Crop classification results on the \textbf{country-specific evaluation}. The best combination of encoder architecture is selected for each fusion strategy. The mean $\pm$ standard deviation between 20 repetitions is shown, scaled between 0 and 100. The  \textcolor{blue}{\textbf{first}} and \textbf{second} best results are highlighted in each country.}\label{tab:bestall:country}
     \footnotesize
    \begin{tabularx}{\linewidth}{LL|L|L|C|C|C}
    \hline
    Country                  & Fusion Strategy                              & Encoder          & {Component}       & AA & Kappa ($\kappa$) & $F_1^{\text{macro}}$ \\ \hline
    \multirow{6}{\linewidth}{Kenya}  &   single-view & TempCNN &  \textit{Radar view} &  $\highest{67.61 \pm 1.19}$ & $\highest{39.68 \pm 3.40}$ & $\highest{68.44 \pm 1.56}$  \\  \cline{2-7} 
    & Input                                      & TAE                   & -     & $\secondhighest{67.25 \pm 5.48}$           & $\secondhighest{37.04 \pm 11.41} $   &  $\secondhighest{67.15 \pm 7.80} $    \\ 
                             &   {Feature}                   & {LSTM} &  -    & ${64.38 \pm 7.24} $  & $28.92 \pm 14.94$ & $62.47 \pm 11.03$  \\ 
                             & {Decision}                  & {GRU}  & G-Fusion        &   $63.73 \pm 6.21$ & ${29.81 \pm 13.67}$ & ${63.59 \pm 7.27}$  \\ 
                             &   Hybrid & {LSTM} & -           & $61.32 \pm 7.04$   & $24.51 \pm 14.98$  & $59.00 \pm 11.39$    \\ 
                             & Ensemble      & LSTM                  & -          & $55.86 \pm 5.88$  & $13.68 \pm 13.49$  & $50.62 \pm 10.08$  \\ \hline 
    \multirow{6}{\linewidth}{Brazil} &    single-view & TAE &  \textit{Optical view} & $97.03 \pm 1.98$ & 	$93.44 \pm 4.94$ &	$96.71 \pm 2.50$   \\  \cline{2-7} 
    & Input                                      & GRU                   & -               &    $95.59 \pm 3.67$  & $92.52 \pm 6.18$    & $96.25 \pm 3.12$       \\ 
                             &   {Feature}       & {LSTM} & -               & $\highest{97.50 \pm 1.43}$  & $\highest{95.77 \pm 2.53}$ & $\highest{97.88 \pm 1.27}$    \\ 
                             & {Decision}      & {LSTM} & Multi-Loss &  $94.91 \pm 7.17$ & $91.67 \pm 12.58$ & $95.73 \pm 6.68$ \\ 
                             &   Hybrid & LSTM                  & -               & $\secondhighest{97.38 \pm 2.35}$ & $\secondhighest{95.64 \pm 3.79}$ & $\secondhighest{97.81 \pm 1.90}$   \\ 
                             & Ensemble        & TempCNN   & -               &   $80.28 \pm 14.17$  & $59.70 \pm 30.41$ & $76.01 \pm 22.27$     \\ \hline 
    \multirow{6}{\linewidth}{Togo}   &   single-view & GRU &  \textit{Optical view} & $80.33 \pm 1.78$ & $55.94 \pm 4.44$ & $77.55 \pm 2.49$ \\  \cline{2-7} 
    & Input                                      & GRU                   & -              & $80.48 \pm 1.42$                  & $56.18 \pm 3.72$ & $77.66 \pm 2.09$  \\ 
                             &   {Feature}        & {GRU}  & -      &   $79.09 \pm 1.35$ & $53.50 \pm 3.17$ & $76.30 \pm 1.79$ \\ 
                             & {Decision}                  & {GRU}  &  Multi-Loss  & $\secondhighest{82.52 \pm 1.58}$ & $\secondhighest{59.82 \pm 3.89}$ & $\secondhighest{79.52 \pm 2.14}$  \\ 
                             &   {Hybrid} & {GRU}  & -  & $80.00 \pm 6.50$   & $55.66 \pm 12.48$  & $76.56 \pm 10.49$   \\ 
                             & Ensemble       & GRU        & -        &  $\highest{84.15 \pm 1.59}$ & $\highest{64.43 \pm 4.35}$ &  $\highest{82.03 \pm 2.32}$  \\ \hline
\end{tabularx}
\end{table*}
Regarding the country-specific evaluation (Table~\ref{tab:bestall:country}), we notice that the Decision fusion is the only strategy that improves its predictions by using additional components (the results with all the components can be found in the Table~\ref{tab_app:country:prediction} in the appendix). 
Additionally, some model combinations produce very unstable results with a high prediction variance. For instance, LSTM encoders with Feature and Hybrid in Kenya, TempCNN encoders with Ensemble in Brazil, and GRU encoders with Hybrid in Togo. This behavior is expected due to the limited amount of data available for training (see Table~\ref{tab:data:samples}). 
Since the best results are achieved with a different model configuration in each country and the high variance in some configurations, we remark that model selection is crucial in these local scenarios with a few labeled samples.

\begin{table*}[!t]
    \centering
    \caption{Crop classification results on the \textbf{global evaluation}. The best combination of encoder architecture is selected for each fusion strategy. The mean $\pm$ standard deviation between 20 repetitions is shown, scaled between 0 and 100. The  \textcolor{blue}{\textbf{first}} and \textbf{second} best results are highlighted in each country.}\label{tab:bestall:global}
    \footnotesize
    \begin{tabularx}{\linewidth}{LL|L|L|C|C|C}
    \hline
    Data                  & Fusion Strategy                              & Encoder          & {Component}       & AA & Kappa ($\kappa$) & $F_1^{\text{macro}}$ \\ \hline
    \multirow{6}{\linewidth}{Global Binary}  &   single-view & TempCNN &  \textit{Optical view} &  $79.98 \pm 0.69$ &   $55.05 \pm 1.30$ & $77.15 \pm 0.69$\\  \cline{2-7} 
    & Input                                      & TempCNN    & -        &    $82.19 \pm 0.73$ & $59.97 \pm 1.61$ & $79.76 \pm 0.86$      \\ 
                             &   {Feature}                   & {TempCNN} &  G-Fusion & $\highest{83.74 \pm 0.53}$  & $\highest{63.64 \pm 0.93}$ & $\highest{81.68 \pm 0.47}$  \\ 
                             & {Decision}                  & {TempCNN}  &  -  & $82.70 \pm 0.61$ & $62.20 \pm 1.03$ & $80.99 \pm 0.52$ \\ 
                             &   Hybrid & {TempCNN} & G-Fusion                & $\secondhighest{83.55 \pm 0.81}$ & $\secondhighest{63.43 \pm 1.55}$ & $\secondhighest{81.58 \pm 0.79}$  \\ 
                             & Ensemble                                   & TempCNN                  & -          & $80.85 \pm 0.58$  & $56.63 \pm 1.08$ & $77.95 \pm 0.57$    \\ \hline 
    \multirow{6}{\linewidth}{Global Multi-Crop} &    single-view & TempCNN &  \textit{Optical view} &  $67.95 \pm 0.70$ &    $60.80 \pm 0.77$ & $59.30 \pm 0.80$   \\  \cline{2-7} 
    & Input                                      & TempCNN                   & -               &   $69.75 \pm 0.96$ & $62.57 \pm 1.10$ & $61.18 \pm 1.31$    \\ 
                             &   {Feature}       & {TempCNN} & G-Fusion &   $\highest{71.11 \pm 0.88}$  & $\secondhighest{64.85 \pm 1.25}$ & $\secondhighest{63.47 \pm 1.38}$  \\ 
                             & {Decision}      & {TempCNN} & -               &  $70.35 \pm 0.74$   & $64.15 \pm 1.35$ & $62.69 \pm 1.44$   \\ 
                             &   Hybrid & TempCNN                  & G-Fusion    & $\secondhighest{70.70 \pm 0.48}$& $\highest{64.99 \pm 0.99}$ & $\highest{63.56 \pm 1.16}$    \\ 
                             & Ensemble            & TempCNN     & -       &      $67.54 \pm 0.54$ & $58.35 \pm 0.92$ & $56.47 \pm 0.93$   \\ \hline 
\end{tabularx}
\end{table*}
In the global evaluation results (Table~\ref{tab:bestall:global}) we observe a clear difference between the multi-class and binary classification. This reflects the challenge of a fine-grained classification involving multiple crops regarding a binary cropland classification.
In these experiments, the G-Fusion component manages to improve the predictions in some fusion strategies, instead of the country-specific evaluation where only with Decision and Kenya data was useful. Besides, the variance in the model results is much lower than in the country-specific, as we expected due to the greater number of labeled training samples. 
The best results in the global evaluation are obtained with the TempCNN encoder for all fusion strategies. We suspect that since the global datasets have more training samples, it is possible to learn the overparameterized TempCNN encoder (Table~\ref{tab:parameters}).
Moreover, the best classification results are obtained with the Feature and Hybrid strategies. Nevertheless, for these scenarios, we recommend using the Feature strategy as it has a smaller number of learnable parameters, and has consistently better results in the country-specific evaluation.

\begin{table*}[!t]
    \centering
    \caption{Crop classification results when selecting the encoder architecture based on the Input fusion strategy. The Togo, and Global evaluation are excluded, since these results can be seen in Table~\ref{tab:bestall:country} and \ref{tab:bestall:global}. The mean $\pm$ standard deviation between 20 repetitions is shown, scaled between 0 and 100. The  \textcolor{blue}{\textbf{first}} and \textbf{second} best results are highlighted in each country.}\label{tab:bestinput}
    \footnotesize
    \begin{tabularx}{\linewidth}{LL|L|L|C|C|C}\hline
    Country                   & Fusion Strategy                               & Encoder           & {Component}    & {AA}        & Kappa ($\kappa$) & {$F_1^{\text{macro}}$} \\ \hline
     \multirow{5}{\linewidth}{Kenya}  & Input                                       &  \multirow{5}{\linewidth}{TAE}                        & -           &   $\highest{67.25 \pm 5.48} $  & $\highest{37.04 \pm 11.41} $   &  $\highest{67.15 \pm 7.80}$  \\ 
                              &     {Feature}        &         &  Multi-Loss    &      $53.80 \pm 5.39$ & $8.85  \pm 11.49$ & $47.00 \pm 8.40$    \\ 
                              &    {Decision}    &   & -   & $53.47 \pm 3.85$ & $8.13 \pm 8.45$  & $47.20 \pm 7.18$\\ 
                             & {Hybrid}  &     &  G-Fusion & $\secondhighest{57.02 \pm 5.69}$ & $\secondhighest{14.77 \pm 11.44}$ & $\secondhighest{53.34 \pm 8.62}$  \\ 
                            & Ensemble      &  & -                                 &     $50.45 \pm 0.53$   & $1.15 \pm 1.30$   & $40.26 \pm 1.46$  \\ \hline 
      \multirow{5}{\linewidth}{Brazil}        & Input      &       \multirow{5}{\linewidth}{GRU}        & -            &   $95.59 \pm 3.67$  & $92.52 \pm 6.18$    & $96.25 \pm 3.12$   \\ 
                              &  {Feature}           &           &  Multi-Loss            &  $\secondhighest{96.57 \pm 2.22}$ & $\highest{94.69 \pm 3.35}$  & $\highest{97.34 \pm 1.68}$  \\ 
                              & {Decision}       &        & - &   $96.46 \pm 4.09$ & $\secondhighest{94.27 \pm 6.52}$    & $\secondhighest{97.13 \pm 3.30}$  \\ 
                              &  {Hybrid}        &     &  -     &  $\highest{96.58 \pm 2.96}$ & $94.22  \pm 4.44$ & $97.10 \pm 2.23$  \\ 
                            & Ensemble      &  & -        &  $70.49 \pm 13.15$  & $34.61 \pm 27.64$ & $59.55 \pm 19.41$ \\ \hline
\end{tabularx}
\end{table*}
Moreover, we explore the question \textit{how much are the predictions conditioned by the fusion strategy given an encoder architecture?} To first select the encoder architecture, we run the five encoder architectures only with the Input fusion and choose the one with the best results. Then, the five fusion strategies are executed with the selected encoder architecture (including the G-Fusion and Multi-Loss variations). This approach reduces the number of experiments compared to the previous one, concretely to 16 experiments ($5 + 5 + 2\cdot3$). The results are shown in Table~\ref{tab:bestinput}. Notice that we do not include the Togo and Global results since those could be observed in Table~\ref{tab:bestall:country} and \ref{tab:bestall:global}, as the fusion strategies already have the same encoder architecture.
Overall, we identify that the other fusion strategies could indeed improve the predictions of the MVL model with Input fusion, except in the Kenya data. 
In the country-specific evaluation, the $\kappa$ score increases from around 2 points in Brazil by the Feature fusion to around 8 points in Togo by the Ensemble strategy. 
The global evaluation shows slight increases in the AA and $F_1^{\text{macro}}$ of 1 to 2 points by the Feature fusion. 
Anyhow, based on the reduction of the number of experiments from the product ($5\cdot 5$) to sum ($5+5$) and the moderate improvement in the classification results, we consider this approach to be worthwhile for future research.

\subsection{Predicted Probabilities Analysis} \label{sec:exp:probabilities}
\begin{table}[!t]
    \centering
    \caption{Predicted probability analysis in the \textbf{country-specific evaluation}. The best configuration is shown for each fusion strategy. The single-view in Kenya is radar, while for Brazil and Togo is optical. The mean $\pm$ standard deviation between 20 repetitions is shown, scaled between 0 and 100.}\label{tab:bestall:country:prob}
    \footnotesize
    \begin{tabularx}{\linewidth}{LL|L|C|C}
    \hline
    Country                  & Strategy                              & Encoder      & Prediction Entropy & Max. Probability \\ \hline
    \multirow{6}{*}{Kenya}  &   single-view & TempCNN &  $42.4 \pm 5.9$ & $88.8 \pm 1.8$  \\  \cline{2-5} 
                            & Input              & TAE      &  $67.1 \pm 7.4$ & $77.7 \pm 3.9$  \\
                             &   {Feature}      & {LSTM}  & $78.5 \pm 8.6$ & $72.2 \pm 5.6$ \\ 
                             & {Decision}           & {GRU}        &  $74.8 \pm 4.0$ & $75.2 \pm 2.2$ \\ 
                             &   Hybrid & {LSTM}  & $91.5 \pm 4.7$ & $63.5 \pm 5.4$ \\ 
                             & Ensemble      & LSTM    &  $85.3 \pm 3.9$ & $69.3 \pm 2.8$ \\ 
                             \hline 
    \multirow{6}{*}{Brazil} &    single-view & TAE   & $66.2 \pm 13.1$ & $78.6 \pm 6.8$ \\  \cline{2-5} 
    & Input                                      & GRU    &   $43.7 \pm 19.1$ & $87.9 \pm 7.8$    \\ 
                             &   {Feature}       & {LSTM}        & $38.5 \pm 16.2$ & $90.3 \pm 5.8$ \\ 
                             & {Decision}      & {LSTM} &  $63.8 \pm 11.8$ & $79.3 \pm 7.7$\\ 
                             &   Hybrid & LSTM  & $86.3 \pm 4.6$ & $68.2 \pm 3.7$  \\ 
                             & Ensemble        & TempCNN   &  $91.9 \pm 4.9$ & $63.0 \pm 5.8$ \\ 
                             \hline 
    \multirow{6}{*}{Togo}   &   single-view & GRU  & $63.3 \pm 3.0$ & $79.3 \pm 1.3$ \\  \cline{2-5} 
    & Input                                      & GRU     & $56.9 \pm 6.0$ & $81.8 \pm 2.4$ \\ 
                             &   {Feature}        & {GRU}  &  $55.6 \pm 4.7$ & $82.2 \pm 1.8$ \\ 
                             & {Decision}                  & {GRU}   &  $71.4 \pm 3.2$ & $76.0 \pm 1.4$  \\ 
                             &   {Hybrid} & {GRU}  & $90.1 \pm 2.3$ & $65.8 \pm 3.0$ \\ 
                             & Ensemble       & GRU       &  $93.7 \pm 0.6$ & $62.4 \pm 0.7$  \\ 
                             \hline
\end{tabularx}
\end{table}
\unskip
As a complementary analysis, we explore the question of \textit{how can multiple input views affect the model's confidence?} To answer this, we compute two measures commonly used to quantify uncertainty \cite{malinin2018predictive}. These only use the prediced probability of models, $\hat{y}^{(i)}_k = \hat{p}(y=k|\mathcal{X}^{(i)})$, i.e. these are independent of the target labels. 
The \textit{max. probability} measuring the model's confidence on the predicted class, $\frac{1}{N}\sum_{i=1}^N \max_k \hat{y}^{(i)}_k$, and the \textit{prediction entropy} measuring the classification uncertainty of the model prediction, $-\frac{1}{N} \sum_{i=1}^N \sum_{k=1}^K \hat{y}^{(i)} \log{ \hat{y}_k^{(i)}}$.
Table~\ref{tab:bestall:country:prob} displays the results in the country-specific evaluation using the best model configuration (in the classification). Results on the global datasets are similar to the ones observed here.
We notice that a model with high max. probability and low prediction entropy obtain the best classification results, such as Feature fusion in Brazil, but not in all cases. Surprisingly, sometimes the less confident and more uncertain predictions are generated by the model with the best classification results, such as the Ensemble strategy in Togo. 
This high uncertainty prediction is associated with the most complex models (Hybrid) or those that do not learn to fuse (Ensemble). 
Moreover, the training may have stopped when a good classification is reached due to early stopping (although the classification is not so confident regarding the predicted probability).
Similar to what is observed in Mena et al. \cite{mena2023crop}, the prediction uncertainty tends to increase slightly as the fusion moves from the input to the output layers of the MVL model, regardless of the encoder architecture.

\begin{figure*}[t!]
\centering
\subfloat[SVL model: Radar with TempCNN encoder.]{\includegraphics[width=0.33\linewidth]{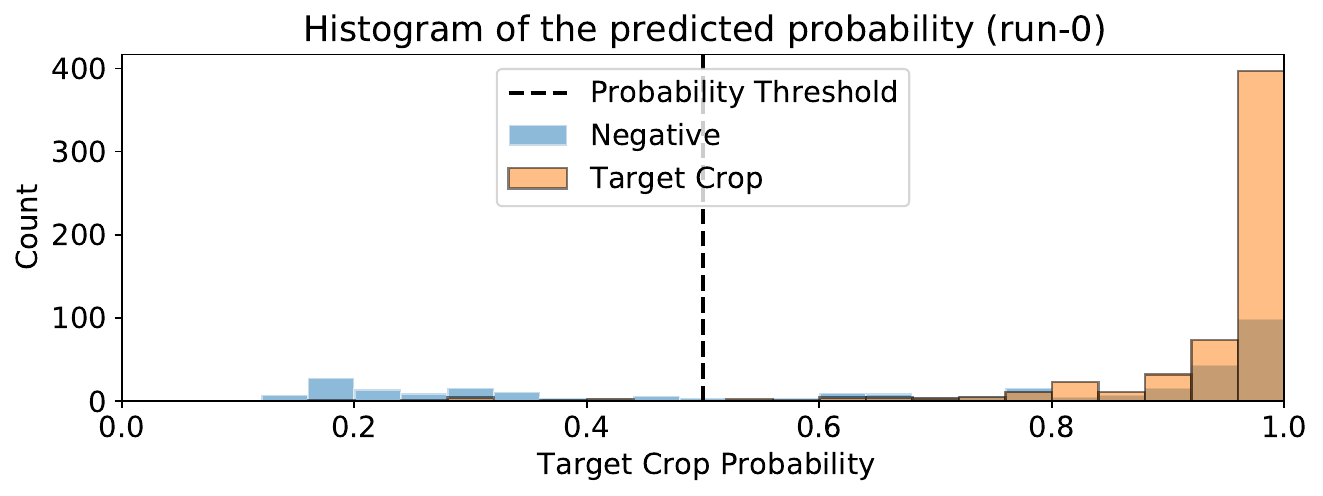}}
\hfill
\subfloat[MVL with low confidence: Hybrid with LSTM.]{\includegraphics[width=0.33\linewidth]{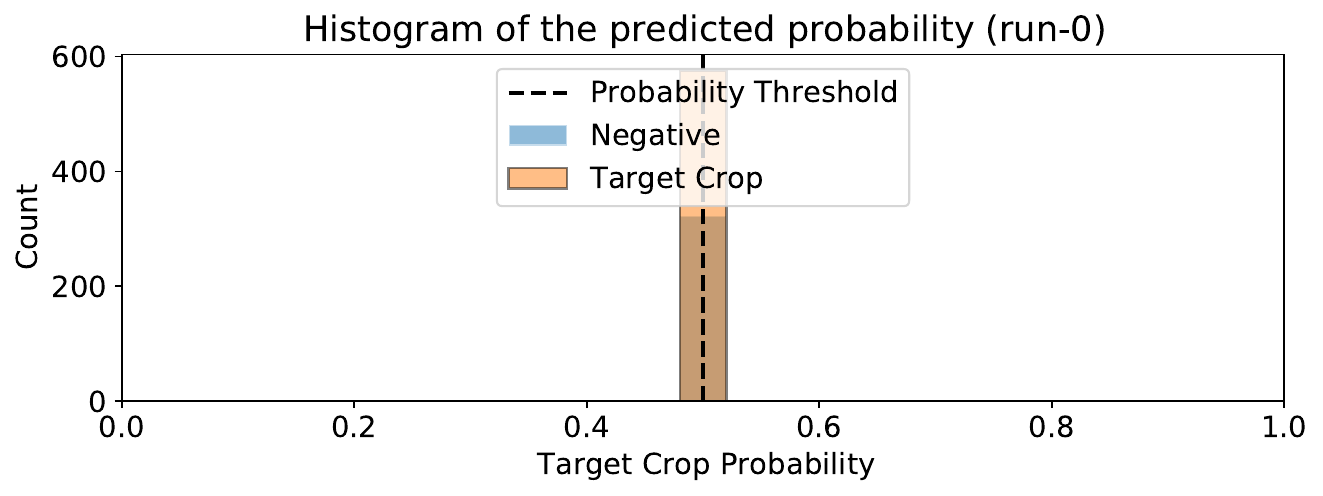}}
\hfill 
\subfloat[MVL with high confidence: Input with TAE.]{\includegraphics[width=0.33\linewidth]{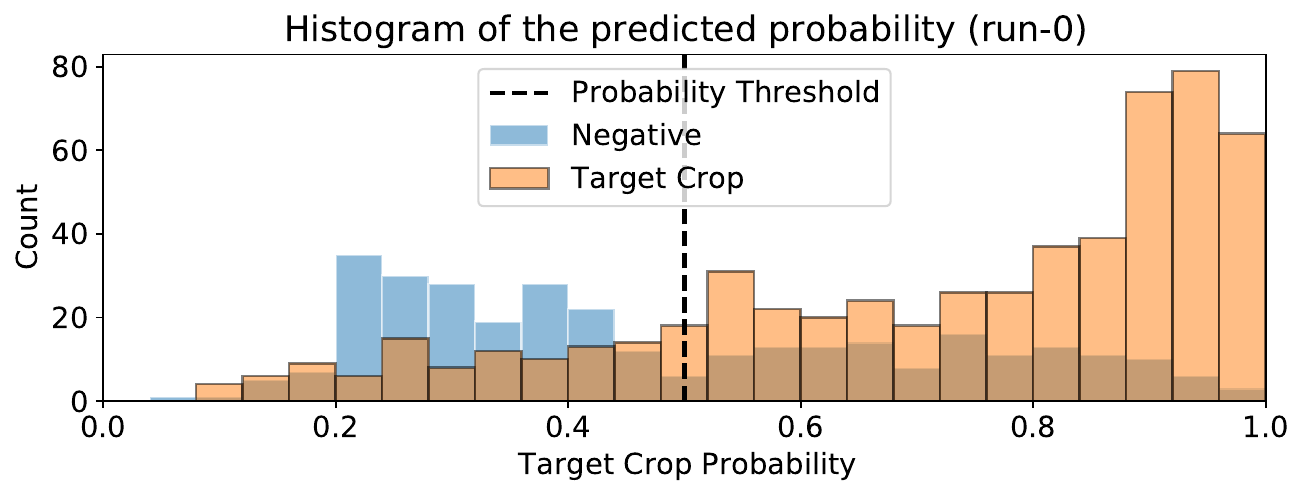}}
\caption{Distribution of the probability predicted by different models in Kenya. The values are taken from a single run.}\label{fig:bestall:kenya:prob}
\end{figure*} 
\begin{figure*}[t!]
\centering
\subfloat[SVL model: Optical with TAE encoder.]{\includegraphics[width=0.33\linewidth]{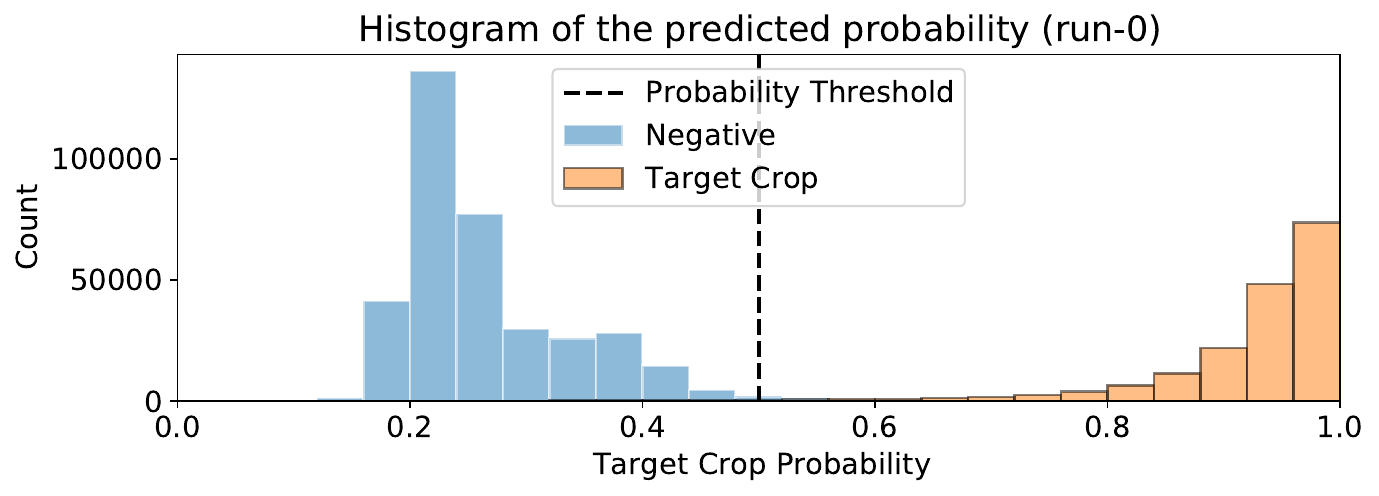}}
\hfill
\subfloat[MVL with low confidence: Ensemble with TempCNN.]{\includegraphics[width=0.33\linewidth]{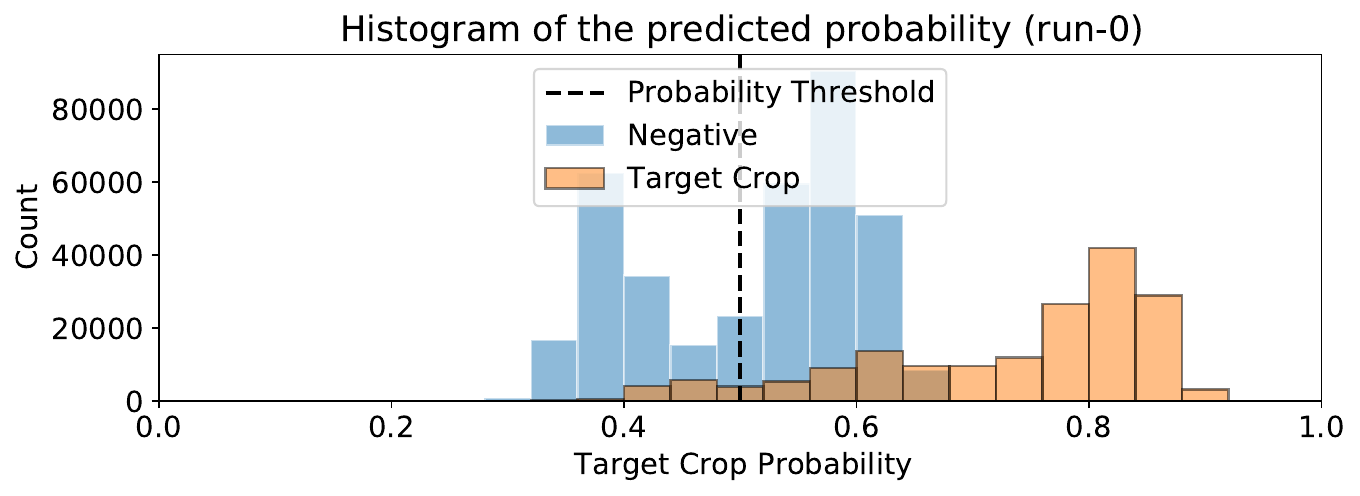}}
\hfill 
\subfloat[MVL with high confidence: Feature with LSTM.]{\includegraphics[width=0.33\linewidth]{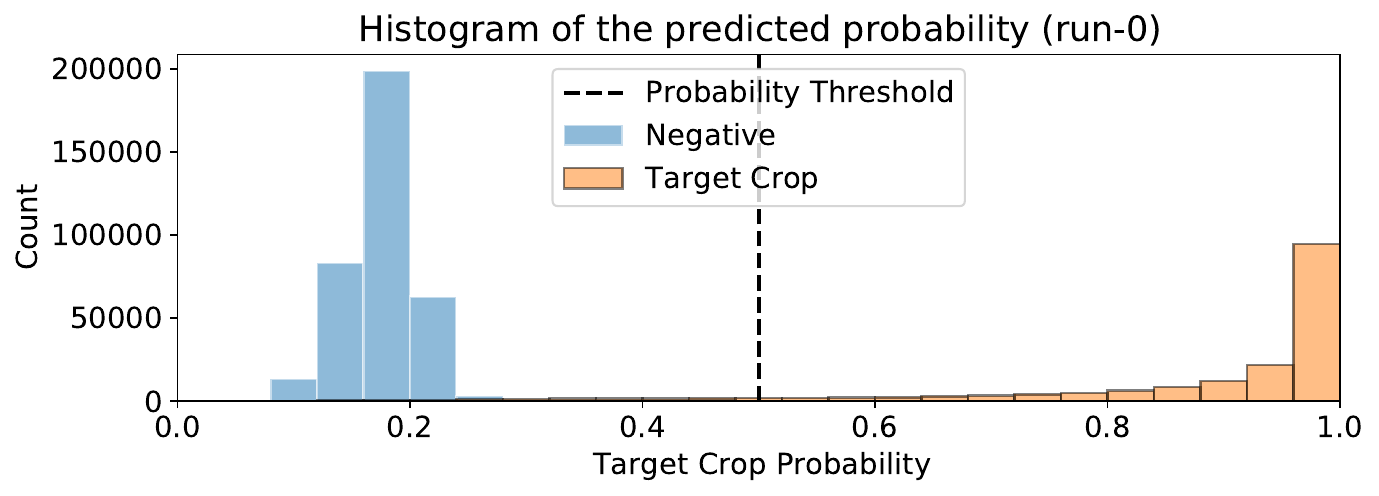}}
\caption{Distribution of the probability predicted by different models in the testing data of Brazil. The values are taken from a single run.}\label{fig:bestall:brazil:prob}
\end{figure*} 
\begin{figure*}[t!]
\centering
\subfloat[SVL model: Optical with GRU encoder.]{\includegraphics[width=0.33\linewidth]{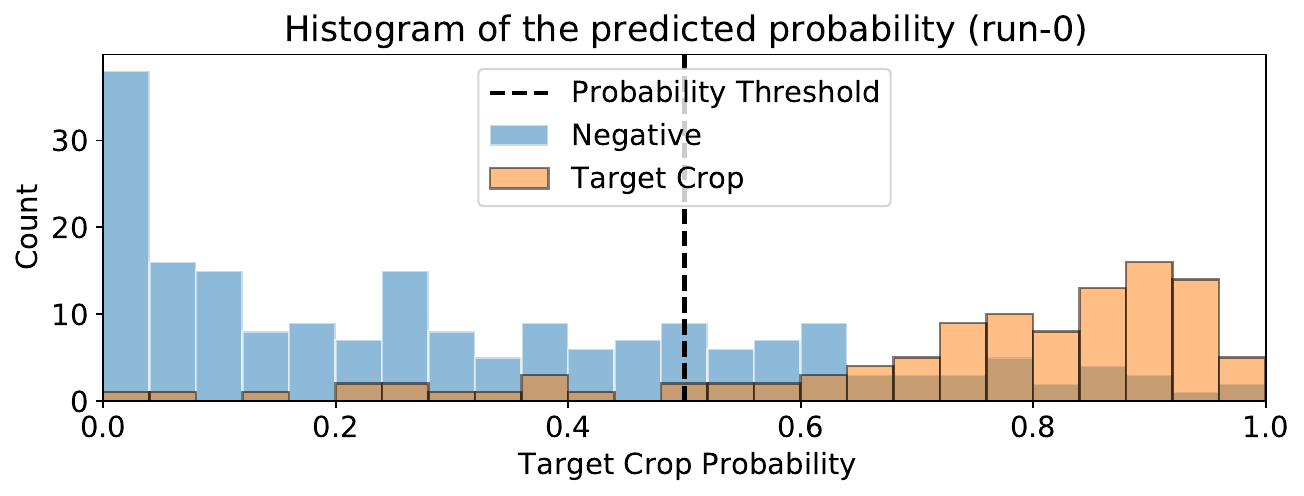}}
\hfill
\subfloat[MVL with low confidence: Ensemble with GRU.]{\includegraphics[width=0.33\linewidth]{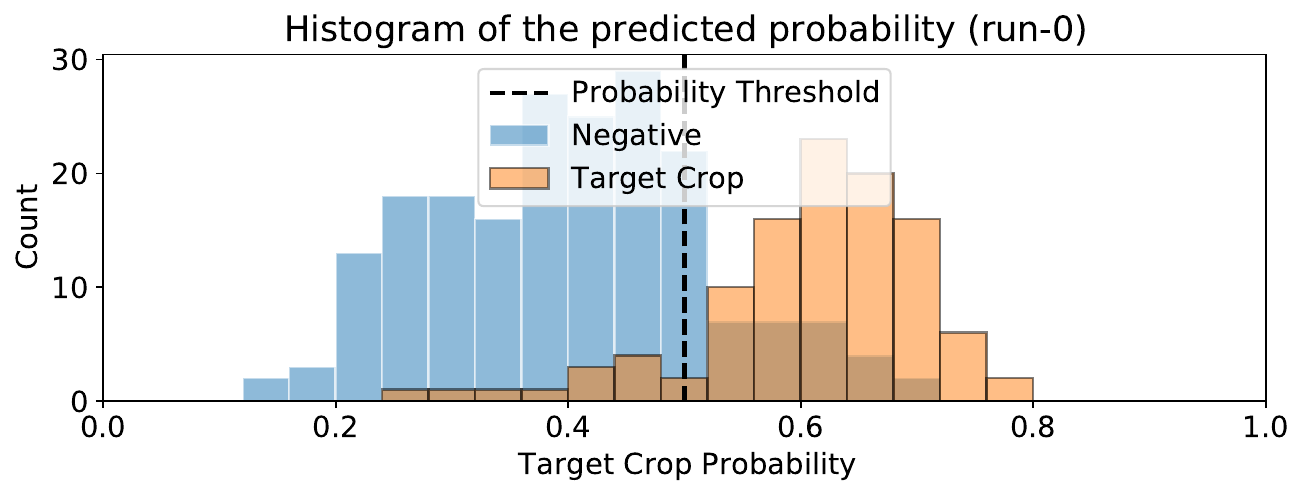}}
\hfill 
\subfloat[MVL with high confidence: Feature with GRU.]{\includegraphics[width=0.33\linewidth]{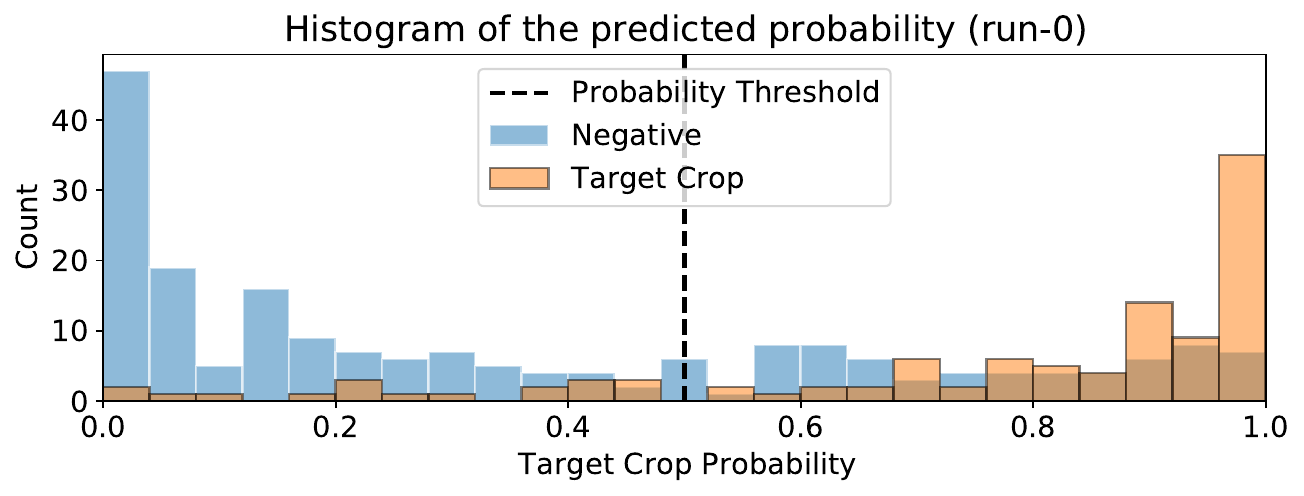}}
\caption{Distribution of the probability predicted by different models in the testing data of Togo. The values are taken from a single run.}\label{fig:bestall:togo:prob}
\end{figure*} 
For a qualitative analysis, we include a chart with the probability distribution associated to the \textit{positive} label in the country-specific evaluation. 
In Fig.~\ref{fig:bestall:kenya:prob}, \ref{fig:bestall:brazil:prob}, and \ref{fig:bestall:togo:prob} we display this for each country in the dataset. We select the best SVL model, an MVL model with confusing predictions, and an MVL model with confident predictions. 
These plots help to interpret the quantities from Table~\ref{tab:bestall:country:prob}. 
Here, a probability prediction closer to the decision boundary of 0.5 (indicating more uncertain predictions) carries a high prediction entropy, while a probability prediction in the extremes (0 or 1, indicating more certain predictions) carries a low prediction entropy.
We can notice how the MVL model assigns more separate values for the different classes, compared to the SVL model, reflecting the advantage of using an MVL methodology.

\subsection{Global Dataset Analysis} \label{sec:exp:further}

We explore the question of \textit{what data patterns can we observe in MVL model predictions?} 
To address this, we provide an analysis of the classification results across different data perspectives, such as crop types, years, and continents.
To extract significant insights, we selected the MVL model with the best classification results in the global evaluation, i.e. the Feature fusion with TempCNN encoders and G-Fusion component.

\begin{table}[!t]
    \centering
    \caption{Crop-wise classification results on the \textbf{Global Multi-crop} evaluation. 
    The mean $\pm$ standard deviation between 20 repetitions is shown, scaled between 0 and 100.}\label{tab:further:crops}
    \footnotesize
    \begin{tabularx}{\linewidth}{l|C|c|C}
    \hline
    Crop-type ($k$)     & $F_{1,k}$   & Precision ($P_k$)        &  Recall ($R_k$) \\ \hline
    Cereals & $83.26 \pm 1.31$ & $77.08 \pm 2.62$ & $90.60 \pm 1.19$ \\
    Others & $75.09 \pm 1.14$  & $66.00 \pm 2.21$ & $87.17 \pm 1.26$ \\ 
    Non-crop & $72.57 \pm 1.13$ & $71.55 \pm 2.55$ & $73.73 \pm 1.87$ \\ 
    Leguminous & $70.32 \pm 2.78$ & $78.86 \pm 4.01$ & $63.63 \pm 3.81$ \\ 
    Oil seeds & $66.07 \pm 3.18$ & $70.62 \pm 3.04$ & $62.28 \pm 4.90$ \\
    Fruits and nuts & $64.81 \pm 2.14$ & $67.22 \pm 4.51$ & $62.81 \pm 2.60$ \\
    Vegetable-melons & $60.40 \pm 1.81$  & $58.78 \pm 3.61$ & $62.31 \pm 2.20$ \\
    Sugar & $58.85 \pm 5.01$ & $84.82 \pm 2.81$ & $45.47 \pm 6.36$ \\
    Root/tuber & $46.48 \pm 3.82$ & $74.84 \pm 3.40$ & $33.97 \pm 4.40$ \\
    Beverage and spice & $36.86 \pm 3.28$ & $61.35 \pm 2.57$ & $26.53 \pm 1.20$ \\
    \hline
\end{tabularx}
\end{table}
Table~\ref{tab:further:crops} displays the $F_1$, precision, and recall scores for each of the ten crop-types in the data. 
It can be seen that the class predictions with the lowest $F_1$ scores are ``beverage and spices'' and ``root/tuber'' crop-types, mainly due to a low recall, i.e. the model fails to identify correctly the samples labeled with those crop-types. This could be explained since these crop-types are among the classes with the least number of training samples (see Fig.~\ref{fig:data:samplesdesc:class}).
On the other hand, the class predictions with higher $F_1$ scores are ``cereals'' and ``others'' crop-types, due mainly to a high recall rather than high precision, i.e. the model identifies correctly most of the samples labeled with those crop-type. 
Similar to the crop-types with the lowest $F_1$ scores, this could be related to the high number of samples in the ``cereals'' and ``others'' crop-types.
This result may be an effect of the model training and the imbalance in the number of samples per crop, as we do not include any balancing step during training.

\begin{table*}[!t]
\begin{minipage}[t]{0.46\textwidth}
    \centering
    \caption{Year-wise classification results on the \textbf{Global Binary} evaluation. 
    The mean $\pm$ standard deviation between 20 repetitions is shown, scaled between 0 and 100.} \label{tab:further:year}
    \footnotesize
    \begin{tabularx}{\linewidth}{l|C|C|C} \hline
        Year &  AA & Kappa ($\kappa$)& $F_1^{\text{macro}}$ \\ \hline
        2016 & $76.97 \pm 0.60$ & $51.01 \pm 1.09$ & $75.06 \pm 0.56$ \\  
        2017 & $85.82 \pm 2.33$ & $72.65 \pm 2.37$ & $86.63 \pm 1.20$ \\
        2018 & $93.78 \pm 0.51$ & $78.25 \pm 0.75$ & $89.08 \pm 0.38$ \\
        2019 & $76.61 \pm 1.72$ & $49.17 \pm 2.00$ & $74.50 \pm 1.02$ \\
        2020 & $89.76 \pm 1.53$ & $81.60 \pm 2.18$ & $90.79 \pm 1.10$ \\
        2021 & $87.51 \pm 0.50$ & $67.21 \pm 1.61$ & $83.33 \pm 0.89$ \\ 
        \hline
    \end{tabularx}
\end{minipage}
\hfill
\begin{minipage}[t]{0.51\textwidth}    
    \centering
    \caption{Continent-wise classification results on the \textbf{Global Binary} evaluation. 
    The mean $\pm$ standard deviation between 20 repetitions is shown, scaled between 0 and 100.
    } \label{tab:further:cont}
    \footnotesize
    \begin{tabularx}{\linewidth}{l|C|C|C} \hline
        Continent &  AA & Kappa ($\kappa$) & $F_1^{\text{macro}}$ \\ \hline
        Africa  & $75.09 \pm 1.20$ & $53.44 \pm 2.22$ & $76.07 \pm 1.24$ \\
        Asia    & $86.68 \pm 0.47$ & $64.97 \pm 1.01$ & $82.33 \pm 0.54$ \\
        Europe  & $79.49 \pm 0.70$ & $54.18 \pm 0.90$ & $76.95 \pm 0.46$ \\
        NA      & $83.26 \pm 1.46$ & $61.98 \pm 1.61$ & $80.92 \pm 0.80$ \\
        Oceania & $77.67 \pm 2.48$ & $52.42 \pm 5.35$ & $75.69 \pm 3.03$ \\
        SA      & $86.87 \pm 1.46$ & $69.34 \pm 1.83$ & $84.61 \pm 0.91$\\ 
        \hline
    \end{tabularx}
\end{minipage}
\end{table*}
In addition, we present the classification results in each continent and year for the Global Binary evaluation. Table~\ref{tab:further:year} presents the AA, $\kappa$, and $F_1^{\text{macro}}$ score for each year, while Table~\ref{tab:further:cont} shows the same results aggregated per continent.
The best AA results are obtained in 2018, which also yields the second-best $\kappa$ and $F_1^{\text{macro}}$. One reason can be that NN models effectively interpolate the training data to learn about the past (pre-2018) and future (post-2018).
Nevertheless, there could be an inverse relation with the number of samples (Table~\ref{tab:data:samples}), since the best $\kappa$ and $F_1^{\text{macro}}$ results are found in 2020, the second year with the least amount of data after 2018. 
In contrast, the worst scores are obtained in 2016 and 2019, the years with the largest number of samples.
Similar behavior occurs in the continent perspective, where the worst AA results are found in Africa, despite it having the second-largest number of samples.
Surprisingly, the classification results are relatively low in Europe, the continent with the largest number of samples.
On the other hand, the best results are in South America and Asia. The top results in South America are interesting, as is the second continent with fewer samples. 

\begin{figure}[!t]
    \centering
    \includegraphics[width=\linewidth]{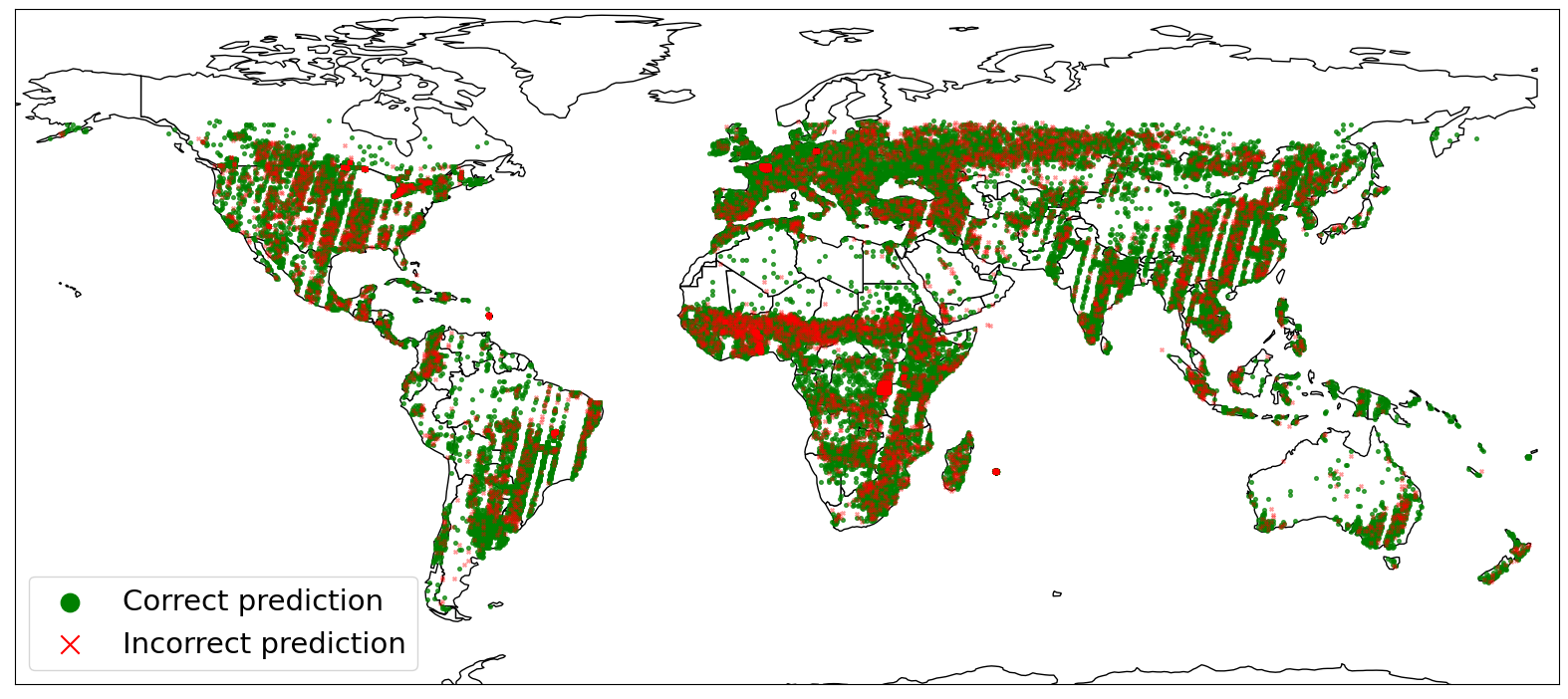}
    \caption{Predicted data points in the Global Binary evaluation. Green circles are points with the predicted class equal to the target class, while red crosses are samples with the predicted class different from the target.}\label{fig:exp:globe}
\end{figure}
Furthermore, to geographically illustrate the mistakes of the MVL model, we display the prediction errors of the training and testing data on a map. To obtain a single prediction of each data point, the best is selected between the 20 repetitions. This is presented in Fig.~\ref{fig:exp:globe}, where we position the errors in the foreground to highlight them. We present the same plots for other methods in the appendix \ref{sec:appendix:fig}, as well as a heatmap error per country.
As depicted in Table~\ref{tab:further:cont}, most of the prediction errors are in Africa, especially in the north of the equator, e.g. in West Africa.
In contrast, there are fewer prediction errors in South America and Asia.
Nevertheless, it can be seen that there are no unique error patterns in localized regions, rather the model makes errors in different zones of the world.

\subsection{State-of-the-art Comparison} \label{sec:exp:sota}

We present a comparison with state-of-the-art results in the CropHarvest dataset \cite{tseng2021crop,tseng2022timl,tseng2023lightweight} in Table~\ref{tab_app:sota}. We display the metrics in the papers focused on the positive class (cropland), with F1 binary and the area under the ROC curve (AUC ROC). We remark that the Presto \cite{tseng2023lightweight} and TIML \cite{tseng2022timl} methods use pre-training strategies, in contrast to our work that only uses supervised training. Besides, for a fair comparison, we show the results from Presto without the dynamic world land cover classes as input.
\begin{table}[!t]
    \centering
    \caption{Crop classification results on the {country-specific evaluation} considering state-of-the-art models. The \textcolor{blue}{\textbf{first}} and \textbf{second} best values are highlighted in each country.}\label{tab_app:sota}
    \footnotesize
    \begin{tabularx}{\linewidth}{ll|C|C}
    \hline
    Country     & Method Name & {F1 \textit{positive}} & {AUC ROC}  \\ \hline
    \multirow{8}{*}{Kenya}  
    & RF \cite{tseng2021crop} &  $55.9 \pm 0.3$  & $57.8 \pm 0.6$  \\ 
    & LSTM Rand \cite{tseng2021crop}  & $78.2 \pm 0.0$ & $32.9 \pm 1.1$  \\ 
    & TIML \cite{tseng2022timl} & $\secondhighest{83.8 \pm 0.0}$ & $\secondhighest{79.4 \pm 0.3}$ \\ 
    & Presto \cite{tseng2023lightweight} & $\highest{86.1 \pm 0.0}$ & $\highest{86.3 \pm 0.0}$ \\ \cline{2-4}
    & Optical-LSTM & $73.6 \pm 4.3$ & $65.1 \pm 4.3$ \\
    & Radar-TempCNN &  $83.2 \pm 0.5$ & $77.6 \pm 2.6$ \\ 
    & Input-TAE & $81.0 \pm 3.2$ & $77.8 \pm  3.7$ \\  
    & Feature-LSTM & $76.1 \pm 5.3$  & $71.6 \pm 6.2$  \\  
    \hline 
    \multirow{8}{*}{Brazil} 
    & RF \cite{tseng2021crop} &  $0.0 \pm 0.0$  & $94.1 \pm 0.4$  \\ 
    & LSTM Rand \cite{tseng2021crop}  & $76.4 \pm 1.2$ & $89.8 \pm 1.0$ \\ 
    & TIML \cite{tseng2022timl} & $83.5 \pm 1.2$ & $98.8 \pm 0.1$ \\ 
    & Presto \cite{tseng2023lightweight} & $88.8 \pm 0.0$ & $98.9 \pm 0.0$   \\ \cline{2-4}
    & Optical-TAE &$95.6 \pm 3.1$ & $\secondhighest{99.5 \pm 0.6}$ \\
    & Radar-GRU & $54.4 \pm 11.2$ &  $80.8 \pm 4.9$ \\ 
    & Feature-LSTM &  $\highest{97.1 \pm 1.7}$ & $\highest{99.6 \pm 0.4}$ \\  
    & Hybrid-LSTM &  $\secondhighest{97.0 \pm 2.7}$ &  $\secondhighest{99.5 \pm 0.3}$ \\ 
    \hline
    \multirow{8}{*}{Togo} 
    & RF \cite{tseng2021crop}  & $75.6 \pm 0.2$  & $89.2 \pm 0.1$ \\ 
    & LSTM Rand \cite{tseng2021crop}   & $72.0 \pm 0.5$ & $86.1 \pm 0.2$ \\ 
    & TIML \cite{tseng2022timl}  & $73.2 \pm 0.2$ & $89.0 \pm 0.0$ \\ 
    & Presto \cite{tseng2023lightweight} & $\secondhighest{76.0 \pm 0.0}$ & $\highest{91.2 \pm 0.0}$ \\ \cline{2-4}
    & Optical-GRU  & $73.6 \pm 2.1$ & $88.6 \pm 0.9$ \\
    & Radar-GRU & $73.5 \pm 2.2$ &  $88.1 \pm 0.4$ \\ 
    & Ensemble-GRU &  $\highest{78.2 \pm 2.2}$ & $\highest{91.2 \pm 0.4}$ \\  
    & Decision-GRU & $\secondhighest{76.0 \pm 1.9}$ & $\secondhighest{90.1 \pm 0.9}$ \\ 
    \hline
\end{tabularx}
\end{table}
In this comparison, we obtained better results in Brazil and a similar AUC in Togo. However, state-of-the-art models are still better in Kenya.
Nevertheless, we remark that our goal is to comprehensively compare different MVL models without relying on pre-training approaches.

\subsection{Time Comparison}
\begin{figure}[!t]
    \centering
    \includegraphics[width=\linewidth]{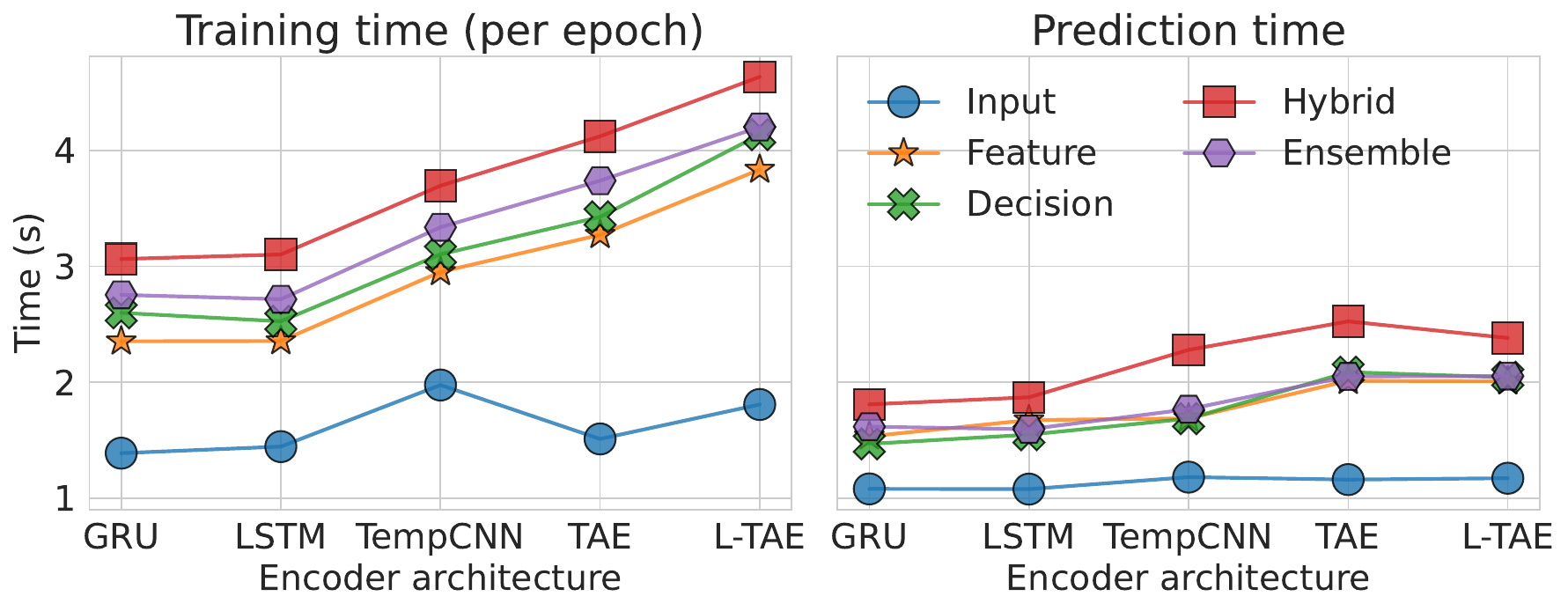}
    \caption{Time comparison between different encoders and fusion strategies.}    \label{fig:time}
\end{figure}
We compare the training and inference time of the different encoders and fusion strategies in Fig.~\ref{fig:time}.
We noticed that the input fusion is twice as fast as the other strategies. 
This is because most of the model parameters come from the encoders. While all other fusions use multiple encoders (and have similar execution times), Input fusion uses only one. 
We note that TempCNN, despite having the highest number of parameters (Table~\ref{tab:parameters}), is not the slowest to execute. It is known in the literature that CNNs have a high capacity for parallelization and acceleration \cite{habib2022optimization}. 
In addition, L-TAE is slower than the TAE encoder despite having fewer parameters. This is because the encoder's optimal configuration allows it to have a larger number of operations.

%% file: content/discussion.tex
\section{Discussion} \label{sec:discussion}

In this section, we address some key points that are relevant to our empirical evidence.

\paragraph{Challenging evaluation scenario in Kenya}
As we present, there are some negative exceptions in the results when looking at Kenya data. Therefore, we wonder \textit{why is there such a different behavior in the Kenya data?} 
In Kenya, the weather view has entropy values dispersed over a broader range of values compared to the narrow values in the other countries (Fig.~\ref{fig:entropy}). Besides, the radar view in Kenya has slightly more entropy than the other views, relative to the other countries. This could explain why the best SVL model is obtained with the radar view instead of the optical (see Table~\ref{tab_app:country:prediction} in the appendix). 
Additionally, Kenya is the country with the highest number of low outliers in the spectral entropy, among the three countries (Fig.~\ref{fig:entropy}), meaning it contains ``irregular'' samples with non-informative patterns. 
These can suggest why this evaluation is more difficult than the others and that the best classification model relies on the highest entropy view (SVL with radar), or a simple fusion strategy with an advanced encoder (Input fusion with TAE encoder).
However, when the SVL model uses the same encoder architecture as the Input Fusion, the TAE encoder, the classification results are improved by MVL models (see Table~\ref{tab_app:input_based} in the appendix). 
This suggests that the better results of the SVL compared to MVL models may only be circumstantial for the model configuration and dataset.
Nevertheless, in other RS-based studies, fewer data sources are somehow better for prediction, such as in crop yield prediction \cite{kang2020comparative,pathak2023predicting}.

\paragraph{Fusion strategy selection based on number of samples}
When training with a high number of labeled samples, such as our global evaluation with more than $19$ thousand samples (see Table~\ref{tab:data:samples}), the predictive quality of the compared MVL models tends to become similar. 
The difference in the classification results between the best fusion strategies is negligible, making the selection of the strategy less critical. 
However, with a limited number of training samples, the choice of the most suitable model has a significant impact on the overall classification result, emphasizing the need for a careful model selection. 
When training with few labeled samples like our local (country-specific) evaluation with less than $1.6$ thousand samples, it is more beneficial to use specialized models that are tailored to the specific data. 
The behavior that the optimal fusion strategy depends on the dataset and region has been observed in both, the RS domain \cite{masolele2021spatial,saintefaregarnot2022multi,mena2023crop} and the computer vision domain \cite{ma2022multimodal}.

\paragraph{Disadvantages of selecting the encoder with the Input fusion}
When the encoder architecture is selected based on the Input fusion, we wonder \textit{how much worse the prediction of the MVL models are by limiting the encoder in this selection?}
In Togo, Global Binary, and Global Multi-crop, the prediction does not get worse since the best results with all fusion strategies are obtained with the same encoder architecture as the Input fusion.
However, in Brazil's evaluation, the quality of the predictions worsens slightly. When using the encoder architecture selected from the Input fusion, the GRU encoder, the results from all fusion strategies are reduced by around $1\%$ compared to the best encoder in each fusion strategy. 
In Kenya's evaluation, the predictive quality worsens significantly. For instance, considering the second-best fusion strategy, the Feature fusion, the AA values are reduced by around $11\%$ from LSTM (the best encoder architecture for Feature) to TAE (the best encoder architecture for Input). Furthermore, the second-best fusion strategy for the $\kappa$ score, the Decision strategy, reduces the values by around $22\%$ from GRU to TAE encoders.
Nevertheless, in most cases the predictions do not vary much and the number of experiments in this approach is significantly reduced.
Therefore, instead of searching for the best encoder architecture for each fusion strategy, an alternative is to find the best encoder for the Input fusion, and then focus on selecting the optimal fusion strategy.

\paragraph{Fusion strategy recommendation for crop classification}
Based on all the experimentation presented in this manuscript, we suggest using Feature fusion, as is the strategy with more points in favor.
The MVL models generated with this strategy obtain either the best or in the middle classification results across evaluation scenarios, without being so complex or so simple (in terms of learnable parameters). Besides, these MVL models generate the most confident predictions.
Moreover, different MVL components could be incorporated based on the flexibility of NN models \cite{mena2023common}, such as different fusion mechanisms (gated fusion or probabilistic fusion), regularization constraints (multiple losses or parameter sharing), or modular design (pre-training encoders or transferring pre-trained layers).
The Input and Ensemble strategies are a good starting point based on the implementation simplicity and competitive results.
However, these strategies are limited to the use of components for individual models, usually related to the architecture design, e.g. number of layers, types of layers, dropout, batch-normalization. 
Our findings and recommendations are further aligned with other results from the literature in crop-related tasks \cite{cuelarosa2018dense,ofori-ampofo2021crop,saintefaregarnot2022multi}, where the Feature fusion obtains either the best or second best classification results.

%% file: content/conclusion.tex
\section{Conclusion} \label{sec:conclusion}

We present an exhaustive comparison of MVL models in a crop classification (cropland and crop-type) by varying the encoder architectures of temporal views and fusion strategy between five options each. 
We assess the pixel-wise classification results in the CropHarvest dataset over various evaluation scenarios.
Our main finding is to corroborate the prediction benefits of using multiple RS sources in a model compared to using just one.
Besides, we find that in specific regions with a limited amount of labeled data, it is better to search for specialized solutions, \textit{one model does not fill all}. 
To find the best encoder architecture and fusion strategy, the search space can be reduced by searching for the best encoder in one fusion strategy first (we use Input fusion), and then varying the fusion strategy.
Furthermore, determining which view or part of the MVL model contributes more to obtaining a better prediction requires further analysis, such as adapting explainability techniques to the MVL scenario.

\paragraph{Limitations}
Although the Cropharvest dataset \cite{tseng2021crop} is a harmonization of 20 public crop datasets (including DENETHOR - AI4EO Food Security and LEM+), the results obtained are conditioned to the configuration of this dataset. For instance, the re-sampling and interpolation to obtain the temporal and spatial resolutions, the labeling harmonization process (between polygons and point data), and crop-type classes.
In our work, we only vary the encoder architecture for the temporal views, while the static view is fixed with an MLP encoder.
Moreover, additional refinements can be expected if a bigger search space is used when looking for architectures.
We are aware of the human bias in the comparison selection, and that recent works, like multi-modal neural architecture search, can search for arbitrary NN architectures using a large amount of computational resources. 
However, our goal is to provide a guided and transparent recommendation for researchers with lower access to computational resources and advance the field of RS-based crop classification.

%% file: content/end_arxiv.tex
\section*{Acknowledgments}
F. Mena acknowledges support through a scholarship of the University of Kaiserslautern-Landau.
We acknowledge H. Najjar, J. Siddamsety, and M. Bugue\~no for providing valuable comments to the manuscript. 

\section*{Data Availability}
The CropHarvest data \cite{tseng2021crop} that is used in this paper can be accessed at \rurl{github.com/nasaharvest/cropharvest}.
Code for the models presented, with the encoders and fusion strategies, is provided at \rurl{github.com/fmenat/optimal-multiview-crop-classifier}.

\section*{Abbreviations}
The following abbreviations are used in this manuscript:\\
\noindent 
\begin{tabularx}{\linewidth}{lL}
CNN & Convolutional Neural Network \\
ERA5 & ECMWF ReAnalysis v5 \\
GRU & Gated Recurrent Unit \\
LSTM & Long-Short Term Memory \\
L-TAE & Lightweight TAE \\
MLP & Multi-Layer Perceptron \\
MVL & Multi-View Learning \\
NDVI & Normalized Difference Vegetation Index \\
NN & Neural Network \\
RF & Random Forest \\
RS & Remote Sensing \\
RNN & Recurrent Neural Network \\
SAR & Synthetic Aperture Radar \\
SRTM & Shuttle Radar Topography Mission\\
SITS & Satellite Image Time Series \\
SVL & Single-View Learning \\
TAE & Temporal Attention Encoder \\
TempCNN & Temporal CNN \\
\end{tabularx}

%% file: content/supp.tex
\subsection{Dataset description}
\begin{figure}[h!]
    \centering
    \includegraphics[width=0.475\textwidth]{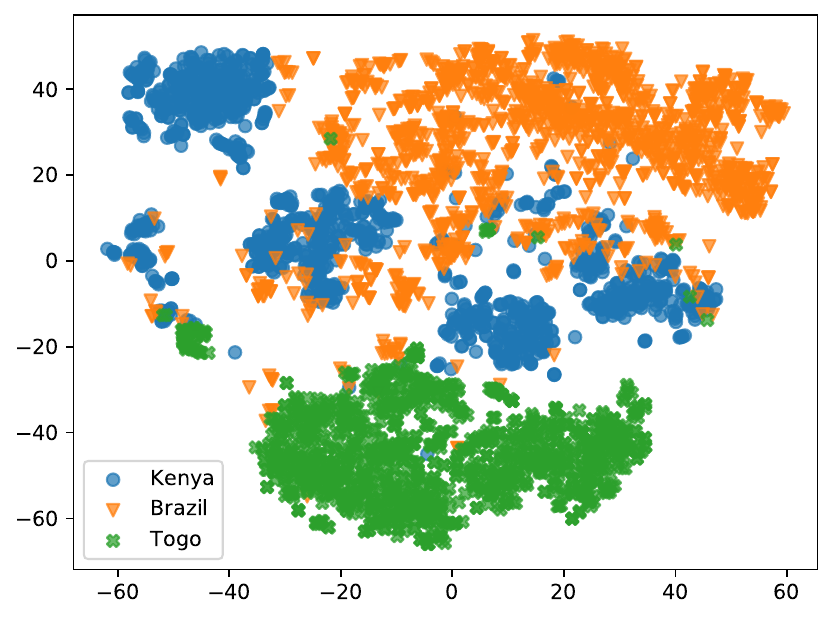}
    \caption{Manifold visualization of the three evaluation countries in the CropHarvest dataset. The t-SNE method is used to generate a 2D projection from the concatenation of the available multi-view input data.}
    \label{fig:data:tsne}
\end{figure}

A visualization of the training data from the three evaluation countries is presented in Fig.~\ref{fig:data:tsne}. Here, the t-distributed Stochastic Neighbor Embedding (t-SNE, \cite{van2008visualizing}) method is used to obtain a projection based on the flattened version of the multi-view data (vector-wise). 
Even though the separation between points from different regions is not perfect, most of the samples from the same region are clustered together, suggesting a region-dependent behavior in the input data.

\subsection{Additional Tables} \label{sec:appendix:tab}

We present a comparison with the state-of-the-art results from Tseng et al. papers \cite{tseng2021crop,tseng2022timl,tseng2023lightweight} in Table~\ref{tab_app:sota}. Here, the RF \cite{tseng2021crop} corresponds to an Input fusion with a RF model, LSTM Rand \cite{tseng2021crop} is an Input fusion with LSTM encoder, TIML \cite{tseng2022timl} is a LSTM Rand with pretext tasks for pre-training, and Presto \cite{tseng2023lightweight} is an Input fusion with transformer encoder and pre-training via reconstruction.
    
We include an extension of the country-specific results from Table~\ref{tab:bestall:country} by including all the component and fusion strategy combinations, shown in Table~\ref{tab_app:country:prediction}. Additionally, we include an extension of the global results from Table~\ref{tab:bestall:global} in the same way, shown in Table~\ref{tab_app:global:prediction}. When analyzing the AUC ROC metric, we notice that the differences in results between models are insignificant, which makes comparison difficult. 
Finally, the same extension is presented from the results of Table~\ref{tab:bestinput}, shown in Table~\ref{tab_app:input_based}.
The same results presented in the main content are observed here. The purpose of these tables is to be a complement to the results for different model configurations.
\begin{table*}[!t]
    \centering
    \caption{Crop classification results on the {country-specific evaluation}, \textbf{extended} from Table~\ref{tab:bestall:country}. The best combination of encoder architecture is selected for each fusion strategy. The mean $\pm$ standard deviation between 20 repetitions is shown, scaled between 0 and 100. The  \textcolor{blue}{\textbf{first}} and \textbf{second} best results are highlighted in each country.}\label{tab_app:country:prediction}
    \small
    \begin{tabularx}{\textwidth}{LL|l|l|C|C|C} \hline
    Country                  & Fusion Strategy                              & View-Encoder          & {Component}       & AA & Kappa & F1 macro \\ \hline
    \multirow{13}{\hsize}{Kenya}  & \multirow{4}{*}{single-view} & LSTM &     \textit{Optical view} & $59.74 \pm 2.51$ & $20.35 \pm 5.39$ & $59.47 \pm 3.20$  \\
     &  & TempCNN &   \textit{Radar view} & $67.61 \pm 1.19$ & $39.68 \pm 3.40$ & $68.44 \pm 1.56 $  \\ 
     & & GRU/LSTM/L-TAE & \textit{Weather view} & $50.00 \pm 0.0$ & $0.00 \pm 0.0$ & $39.04 \pm 0.0$ \\ 
     & & MLP & \textit{Topography view} & $49.30 \pm 0.69$ & $-1.77 \pm 1.74$ & $39.00 \pm 1.20$ \\
     \cline{2-7} 
    & Input                                      & TAE                   & -                                    & $\highest{67.25 \pm 5.48}$           & $\highest{37.04 \pm 11.41} $   &  $\highest{67.15 \pm 7.80} $   \\ \cline{2-7} 
                             & \multirow{3}{*}{Feature}                   & \multirow{3}{*}{LSTM} &  -                                    & $\secondhighest{64.38 \pm 7.24} $  & $28.92 \pm 14.94$ & $62.47 \pm 11.03$  \\ 
                             &                                            &                       &    Multi-Loss                                  & $62.52 \pm 8.91$ & $25.43 \pm 18.49$ & $57.93 \pm 16.24$  \\ 
                             &                                            &                      &  G-Fusion                                     &    $58.84 \pm 7.34$ & $18.24 \pm 15.04$ & $55.31 \pm 11.96$  \\ \cline{2-7} 
                             & \multirow{3}*{Decision}                  & \multirow{3}{*}{GRU}  & -                                    & $58.66 \pm 6.94$ & $18.55 \pm 15.22$ & $58.33 \pm 7.63$ \\ 
                             &                                            &             &        Multi-Loss                                    &     $62.28 \pm 5.88$ & $27.06 \pm 12.57$ & $61.64 \pm 7.57$     \\ 
                             &                                            &                       & G-Fusion                                     &          $63.73 \pm 6.21$ & $\secondhighest{29.81 \pm 13.67}$ & $\secondhighest{63.59 \pm 7.27}$  \\      \cline{2-7} 
                             & \multirow{3}{*}{Hybrid} & \multirow{3}{*}{LSTM} & -                                    & $61.32 \pm 7.04$   & $24.51 \pm 14.98$  & $59.00 \pm 11.39$                      \\ 
                             &                                            &                       &   Multi-Loss                                   & $59.55 \pm 6.53$ & $20.35 \pm 13.72$ & $56.74 \pm 11.16$  \\ 
                             &                                            &                       &  G-Fusion                                      &  $59.87 \pm 9.57$ & $18.22 \pm 17.63$ & $53.10 \pm 14.56$ \\ \cline{2-7} 
                             & Ensemble                                   & LSTM                  & -                                    & $55.86 \pm 5.88$  & $13.68 \pm 13.49$  & $50.62 \pm 10.08$    \\ \hline 
    \multirow{13}{\hsize}{Brazil} &  \multirow{4}{*}{single-view} & TAE &  \textit{Optical view} & $97.03 \pm 1.98$ & 	$93.44 \pm 4.94$ &	$96.71 \pm 2.50$  \\  
    &  & GRU &  \textit{Radar view} &  $64.57 \pm 6.77$ & $26.64 \pm 15.52$ & $58.27 \pm 11.74$  \\ 
    & & GRU & \textit{Weather view} & $50.00 \pm 0.0 $ & $0.00 \pm 0.0$ & $25.25 \pm 3.55$ \\ 
    & & MLP & \textit{Topography view} & $50.10 \pm 0.72$ & $0.14 \pm 1.04$ & $25.38 \pm 3.14$ \\
    \cline{2-7} 
    & Input                                      & GRU                   & -               & $95.59 \pm 3.67$  & $92.52 \pm 6.18$    & $96.25 \pm 3.12$         \\ \cline{2-7} 
                             & \multirow{3}{*}{Feature}                   & \multirow{3}{*}{LSTM} &  -               & $\highest{97.50 \pm 1.43}$  & $\highest{95.77 \pm 2.53}$ & $\highest{97.88 \pm 1.27}$     \\
                             &                                            &                       & Multi-Loss &   $94.56 \pm 10.88$ & $90.29 \pm 21.71$ & $93.86 \pm 16.49$        \\  
                             &                                            &                       &  G-Fusion    &  $88.27 \pm 16.77$ & $77.89 \pm 33.94$ & $85.90 \pm 24.51$   \\ \cline{2-7} 
                             & \multirow{3}{*}{Decision}                  & \multirow{3}{*}{LSTM} & -               & $94.15 \pm 5.20$ & $90.73 \pm 8.30$  & $95.33 \pm 4.21$     \\ 
                             &                                            &                       &  Multi-Loss &  $94.91 \pm 7.17$ & $91.67 \pm 12.58$ & $95.73 \pm 6.68$   \\ 
                             &                                            &                       & G-Fusion    &   $91.59 \pm 3.93$ & $86.44 \pm 6.33$ & $93.18 \pm 3.21$    \\ \cline{2-7} 
                             & \multirow{3}{*}{Hybrid} & \multirow{3}{*}{LSTM}                  & -               & $\secondhighest{97.38 \pm 2.35}$ & $\secondhighest{95.64 \pm 3.79}$ & $\secondhighest{97.81 \pm 1.90}$      \\ \
                             &                                            &                       & Multi-Loss &  $86.16 \pm 18.72$ & $74.09 \pm 38.20$ & $83.52 \pm 26.55$   \\  
                             &                                            &                       &  G-Fusion    &   $70.73 \pm 18.35$ & $44.81 \pm 37.58$ & $67.11 \pm 25.05$    \\ \cline{2-7} 
                             & Ensemble                                   & TempCNN               & -               & $80.28 \pm 14.17$  & $59.70 \pm 30.41$ & $76.01 \pm 22.27$ \\ \hline 
    \multirow{13}{\hsize}{Togo}   &  \multirow{4}{*}{single-view} & GRU &   \textit{Optical view} & $80.33 \pm 1.78$ & $55.94 \pm 4.44$ & $77.55 \pm 2.49$ \\ 
    & & GRU &  \textit{Radar view} & $79.97 \pm 1.94$ & $56.96 \pm 5.33$ & $78.17 \pm 3.13$ \\     
    & & TempCNN & \textit{Weather view} & $57.89 \pm 2.95$ & $15.14 \pm 5.47$ & $57.01 \pm 3.19$ \\ 
    & & MLP & \textit{Topography view} & $61.31 \pm 2.16$ & $20.54 \pm 3.66$ & $59.32 \pm 1.77$ \\ 
    \cline{2-7} 
    & Input                                      & GRU                   & -                                    & $80.48 \pm 1.42$                  & $56.18 \pm 3.72$ & $77.66 \pm 2.09$  \\ \cline{2-7} 
                             & \multirow{3}{*}{Feature}                   & \multirow{3}{*}{GRU}  &  -                                    & $79.09 \pm 1.35$ & $53.50 \pm 3.17$ & $76.30 \pm 1.79$  \\
                             &                                            &                       &  Multi-Loss                                     &  $77.04 \pm 5.53$ & $49.58 \pm 10.72$ & $73.73 \pm 7.14$    \\ 
                             &                                            &                       &   G-Fusion                                    &         $77.19 \pm 3.27$ & $49.27 \pm 6.96$ & $73.79 \pm 4.60$   \\ \cline{2-7} 
                             & \multirow{3}{*}{Decision}                  & \multirow{3}{*}{GRU}  &  -                                    & $82.24 \pm 1.33$ & $59.74 \pm 3.66$ & $\secondhighest{79.53 \pm 2.05}$  \\
                             &                                            &                       &  Multi-Loss                   & $\secondhighest{82.52 \pm 1.58}$ & $\secondhighest{59.82 \pm 3.89}$ & $79.52 \pm 2.14$    \\  
                             &                                            &                       &    G-Fusion                                   &  $80.43 \pm 2.74$ & $56.52 \pm 6.63$ & $77.82 \pm 3.80$   \\ \cline{2-7} 
                             & \multirow{3}{*}{Hybrid} & \multirow{3}{*}{GRU}  &  -                                    & $80.00 \pm 6.50$   & $55.66 \pm 12.48$  & $76.56 \pm 10.49$   \\ 
                             &                                            &                       &   Multi-Loss                                    &    $79.64 \pm 6.62$ & $54.66 \pm 12.57$ & $75.94 \pm 10.82$ \\ 
                             &                                            &                       &    G-Fusion                                   &  $76.53 \pm 9.45$ & $48.81 \pm 17.48$ & $71.74 \pm 15.41$   \\ \cline{2-7} 
                             & Ensemble                                   & GRU                   & -                                    & $\highest{84.15 \pm 1.59}$ & $\highest{64.43 \pm 4.35}$ & $\highest{82.03 \pm 2.32}$  \\ \hline
\end{tabularx}
\end{table*}

\begin{table*}[!t]
    \centering
    \caption{Crop classification results on the {global evaluation}, \textbf{extended} from Table~\ref{tab:bestall:global}. The best combination of encoder architecture is selected for each fusion strategy. The mean $\pm$ standard deviation between 20 repetitions is shown, scaled between 0 and 100. The  \textcolor{blue}{\textbf{first}} and \textbf{second} best results are highlighted in each country.}\label{tab_app:global:prediction}
    \small
    \begin{tabularx}{\linewidth}{LL|l|l|C|C|C}
    \hline
    Data     & Fusion Strategy    & View-Encoder & Component    & {AA} & {Kappa} & {F1 macro}  \\ \hline
    \multirow{13}{\hsize}{Global Binary}     & \multirow{4}{*}{single-view} & TempCNN  &  \textit{Optical view} & $79.98 \pm 0.69$ &	$55.05 \pm 1.30$ & $77.15 \pm 0.69$  \\ 
    &  & TempCNN & \textit{Radar view}  & $74.35 \pm 0.93$ & $44.01 \pm 1.89$ & $71.34 \pm 1.11$ \\  
    & & TempCNN & \textit{Weather view} & $74.79 \pm 0.60$ & $46.03 \pm 1.69$ & $72.65 \pm 1.00$ \\
    & & MLP & \textit{Topography view} & $64.96 \pm 0.09$ & $28.53 \pm 0.30$ & $64.08 \pm 0.23$ \\
    \cline{2-7}
    & Input                                      & TempCNN      & -          & $82.19 \pm 0.73$ & $59.97 \pm 1.61$ & $79.76 \pm 0.86$                     \\ \cline{2-7} 
                                        & \multirow{3}{*}{Feature}                   & \multirow{3}{*}{TempCNN}      &  -              & $83.13 \pm 1.03$ & $62.55 \pm 1.72$ & $81.14 \pm 0.85$  \\ 
                                        &                                            &      & Multi-Loss    & $\secondhighest{83.71 \pm 0.44}$ & $\highest{63.65 \pm 1.03}$ & $\highest{81.69 \pm 0.54}$  \\ 
                                        &                                          &      &  G-Fusion     & $\highest{83.74 \pm 0.53}$  & $\secondhighest{63.64 \pm 0.93}$ & $\secondhighest{81.68 \pm 0.47}$  \\ \cline{2-7} 
                                        & \multirow{3}{*}{Decision}                  & \multirow{3}{*}{TempCNN}      & -      & $82.70 \pm 0.61$ & $62.20 \pm 1.03$ & $80.99 \pm 0.52$ \\
                                        &                                            &       &  Multi-Loss      & $82.20 \pm 0.71$  & $60.18 \pm 1.51$ & $79.88 \pm 0.80$ \\ 
                                        &                                            &      & G-Fusion  & $82.64 \pm 0.79$ & $61.59 \pm 1.55$ & $80.65 \pm 0.79$  \\ \cline{2-7} 
                                        & \multirow{3}{*}{Hybrid} & \multirow{3}{*}{TempCNN}      &  -              & $\secondhighest{83.71 \pm 0.53}$ & $63.27 \pm 1.11$   & $81.47 \pm 0.58$   \\ 
                                        &       &       & Multi-Loss   & $83.18 \pm 0.89$ & $62.30 \pm 1.66$ & $80.98 \pm 0.85$  \\ 
                                        &                                            &       &  G-Fusion   & $83.55 \pm 0.81$ & $63.43 \pm 1.55$ & $81.58 \pm 0.79$  \\ \cline{2-7} 
                                        & Ensemble                                   & TempCNN      & -    & $80.85 \pm 0.58$  & $56.63 \pm 1.08$ & $77.95 \pm 0.57$  \\ \hline 
    \multirow{13}{\hsize}{Global Multi-Crop} & \multirow{4}{*}{single-view} & TempCNN &  \textit{Optical view} & $67.95 \pm 0.70$ &	$60.80 \pm 0.77$ & $59.30 \pm 0.80$  \\ 
    & & TempCNN &  \textit{Radar view} & $59.39 \pm 0.61$ & $50.33 \pm 1.20$ & $50.27 \pm 1.22$  \\     
    & & TempCNN & \textit{Weather view} & $45.50 \pm 0.71$ & $29.54 \pm 1.44$ & $32.97 \pm 1.14$ \\
    & & MLP & \textit{Topography view} & $29.96 \pm 0.51$ & $15.06 \pm 0.55$ & $19.01 \pm 0.73$ \\ 
    \cline{2-7}
    & Input                                      & TempCNN      & -      & $69.75 \pm 0.96$ & $62.57 \pm 1.10$ & $61.18 \pm 1.31$  \\ \cline{2-7} 
                                        & \multirow{3}{*}{Feature}                   & \multirow{3}{*}{TempCNN}      &  -      & $70.58 \pm 0.71$   & $64.27 \pm 1.31$ & $62.85 \pm 1.36$    \\ 
                                        &                                            &      & Multi-Loss    & $\secondhighest{70.89 \pm 0.59}$  & $64.34 \pm 0.87$  & $62.93 \pm 0.97$   \\  
                                        &                                            &      &  G-Fusion      &  $\highest{71.11 \pm 0.88}$  & $\secondhighest{64.85 \pm 1.25}$ & $\secondhighest{63.47 \pm 1.38}$  \\ \cline{2-7} 
                                        & \multirow{3}{*}{Decision}                  & \multirow{3}{*}{TempCNN}      & -      & $70.35 \pm 0.74$   & $64.15 \pm 1.35$ & $62.69 \pm 1.44$   \\ 
                                        &                                            &       &  Multi-Loss  & $70.47 \pm 0.81$                 & $63.84 \pm 1.51$  & $62.33 \pm 1.63$\\ 
                                        &                                            &       & G-Fusion   & $68.66 \pm 1.18$ & $61.61 \pm 1.92$ & $60.70 \pm 1.97$  \\ \cline{2-7} 
                                        & \multirow{3}{*}{Hybrid} & \multirow{3}{*}{TempCNN}      &  -          & $70.92 \pm 0.62$ & $64.41 \pm 1.20$ & $62.76 \pm 1.49$        \\
                                        &                                            &      & Multi-Loss   & $70.44 \pm 0.69$  & $63.89 \pm 1.01$ & $62.02 \pm 1.13$  \\ 
                                        &                                            &      &  G-Fusion    & $70.70 \pm 0.48$& $\highest{64.99 \pm 0.99}$ & $\highest{63.56 \pm 1.16}$   \\ \cline{2-7} 
                                        & Ensemble                                   & TempCNN      & -    & $67.54 \pm 0.54$ & $58.35 \pm 0.92$ & $56.47 \pm 0.93$ \\ \hline
\end{tabularx}
\end{table*}

\begin{table*}[!t]
    \centering
    \caption{Crop classification results when selecting the encoder architecture based on the Input fusion, \textbf{extended} from Table~\ref{tab:bestinput}.  The mean $\pm$ standard deviation between 20 repetitions is shown, scaled between 0 and 100. The  \textcolor{blue}{\textbf{first}} and \textbf{second} best results are highlighted in each country.}\label{tab_app:input_based}
    \small
    \begin{tabularx}{\linewidth}{LL|L|L|C|C|C}\hline
    Country                   & Fusion Strategy                               & View-Encoder           & {Component}    & {AA}        & {Kappa} & {F1 macro} \\ \hline
     \multirow{13}{\hsize}{Kenya}  & \multirow{2}{\hsize}{single-view} &   &  \textit{Optical view} & $53.51 \pm 3.15$ & $8.23 \pm 6.76$ & $47.3 \pm 6.21$ \\ 
     & & & \textit{Radar view} & $59.59 \pm 7.15$ & $18.48 \pm 13.94$ & $58.10 \pm 8.05$  \\ \cline{2-2} \cline{4-7} 
     & Input                                       &  \multirow{13}{\hsize}{TAE}                        & -       & $\highest{67.25 \pm 5.48} $  & $\highest{37.04 \pm 11.41} $   &  $\highest{67.15 \pm 7.80}$   \\ \cline{2-2} \cline{4-7} 
                              &     \multirow{3}{\hsize}{Feature}     &        &  -        & $52.15 \pm 1.93$   & $5.08 \pm 4.41$ & $45.31 \pm 5.47$     \\ 
                              &                                             &                        & Multi-Loss                        & $53.80 \pm 5.39$ & $8.85  \pm 11.49$ & $47.00 \pm 8.40$   \\
                              &               &                        &  G-Fusion                      & $52.61 \pm 2.23$   & $6.26 \pm 5.24$ & $46.12 \pm 5.22$ \\ \cline{2-2} \cline{4-7} 
                              &    \multirow{3}{\hsize}{Decision}     &     & -     & $53.47 \pm 3.85$ & $8.13 \pm 8.45$  & $47.20 \pm 7.18$  \\
                              &                                             &                        &  Multi-Loss                        & $51.45 \pm 1.13$ & $3.58 \pm 2.70$ & $43.56 \pm 3.21$    \\
                              &       &       & G-Fusion      & $53.38 \pm 4.41$ & $7.77 \pm 9.18$ & $46.70 \pm 7.46$   \\ \cline{2-2} \cline{4-7} 
                             & \multirow{3}{\hsize}{Hybrid}  &       &  -           & $53.91 \pm 2.83$  & $9.19 \pm 6.18$ & $48.17 \pm 6.08$        \\ \
                              &                       &          & Multi-Loss          & $52.74 \pm 2.68$  & $6.48 \pm 5.91$ & $46.22 \pm 5.66$  \\  
                              &  &            &  G-Fusion    & $\secondhighest{57.02 \pm 5.69}$ & $\secondhighest{14.77 \pm 11.44}$ & $\secondhighest{53.34 \pm 8.62}$     \\ \cline{2-2} \cline{4-7} 
      & Ensemble                                    &  & -                                 & $50.45 \pm 0.53$   & $1.15 \pm 1.30$   & $40.26 \pm 1.46$   \\ \hline 
      \multirow{13}{\hsize}{Brazil}        & \multirow{2}{\hsize}{single-view} &   &  \textit{Optical view} &  $93.8 \pm 5.17$ & $89.4 \pm 9.35$ & $94.7 \pm 4.79$  \\
      & & & \textit{Radar view} & $64.61 \pm 6.8$ & $26.72 \pm 15.5$ & $58.27 \pm 11.74$ \\\cline{2-2} \cline{4-7} 
      & Input                                       &       \multirow{13}{\hsize}{GRU}                 & -            & $95.59 \pm 3.67$  & $92.52 \pm 6.18$    & $96.25 \pm 3.12$      \\ \cline{2-2} \cline{4-7} 
                              &      \multirow{3}{\hsize}{Feature}       &       &  -      & $94.03 \pm 2.27$                         & $90.53 \pm 3.60$ & $95.24 \pm 1.80$    \\ 
                              &                 &          & {Multi-Loss}   & $96.57 \pm 2.22$ & $\highest{94.69 \pm 3.35}$  & $\highest{97.34 \pm 1.68}$             \\ 
                              &        &         &  {G-Fusion} & $92.43 \pm 14.68$ & $87.27 \pm 29.22$ & $93.60 \pm 19.97$    \\ \cline{2-2} \cline{4-7} 
                              &   \multirow{3}{\hsize}{Decision}         &         & -       & $96.46 \pm 4.09$ & $\secondhighest{94.27 \pm 6.52}$    & $\secondhighest{97.13 \pm 3.30}$   \\ 
                              &           &         &  {Multi-Loss}   & $\secondhighest{96.12 \pm 2.29}$ & $94.00 \pm 3.48$ & $97.00 \pm 1.75$   \\ 
                              &               &        & {G-Fusion} & $89.59 \pm 4.58$ & $81.09 \pm 7.22$  & $88.74 \pm 3.68$   \\ \cline{2-2} \cline{4-7} 
                              &          \multirow{3}{\hsize}{Hybrid}                                   &                        &  -            & $\highest{96.58 \pm 2.96}$ & $94.22  \pm 4.44$ & $97.10 \pm 2.23$   \\
                              &            &        & {Multi-Loss}   & $96.26 \pm 2.77$ & $94.14 \pm 4.20$ & $97.06 \pm 2.11$  \\ 
                              &  &                        &  {G-Fusion} & $69.12 \pm 17.99$  & $42.42 \pm 37.06$ & $66.53 \pm 24.17$   \\ \cline{2-2} \cline{4-7} 
                            & Ensemble     &  & -   & $70.49 \pm 13.15$  & $34.61 \pm 27.64$ & $59.55 \pm 19.41$  \\ \hline
\end{tabularx}
\end{table*}

\subsection{Additional Figures} \label{sec:appendix:fig}
\begin{figure*}[!t]
    \centering
    \subfloat[Single-view model fed with the optical view.]{\includegraphics[width=0.495\linewidth]{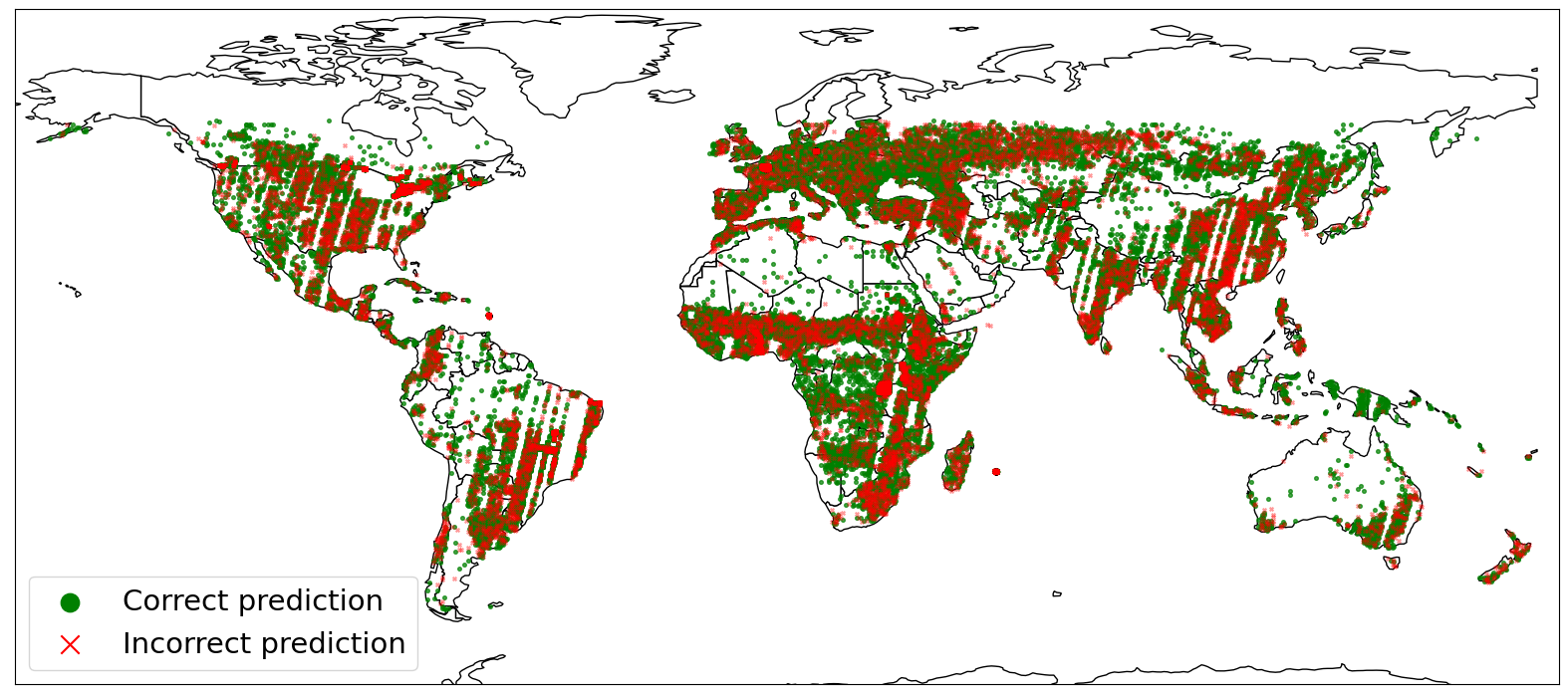}}
    \hfill
    \subfloat[MVL model with Input strategy.]{\includegraphics[width=0.495\linewidth]{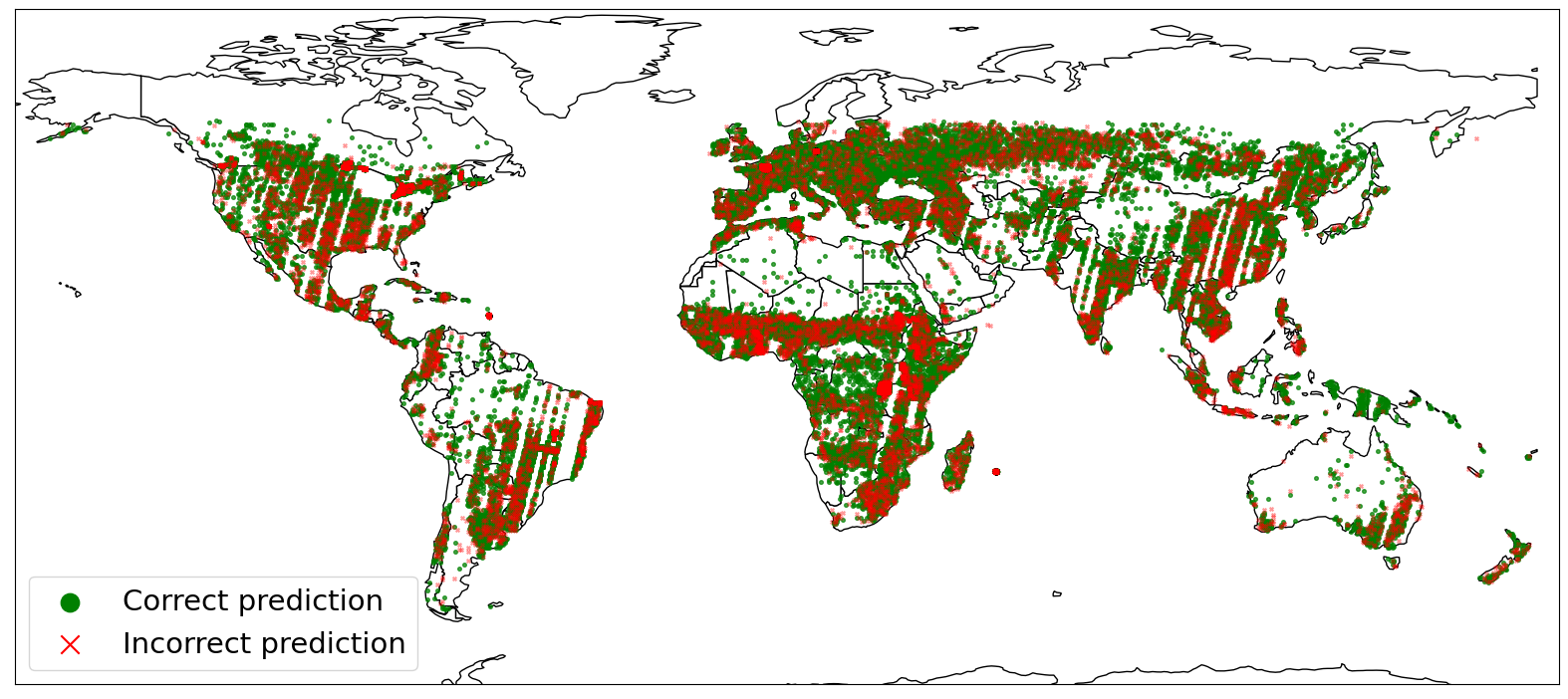}} \\\
    \subfloat[MVL model with Decision strategy.]{\includegraphics[width=0.495\linewidth]{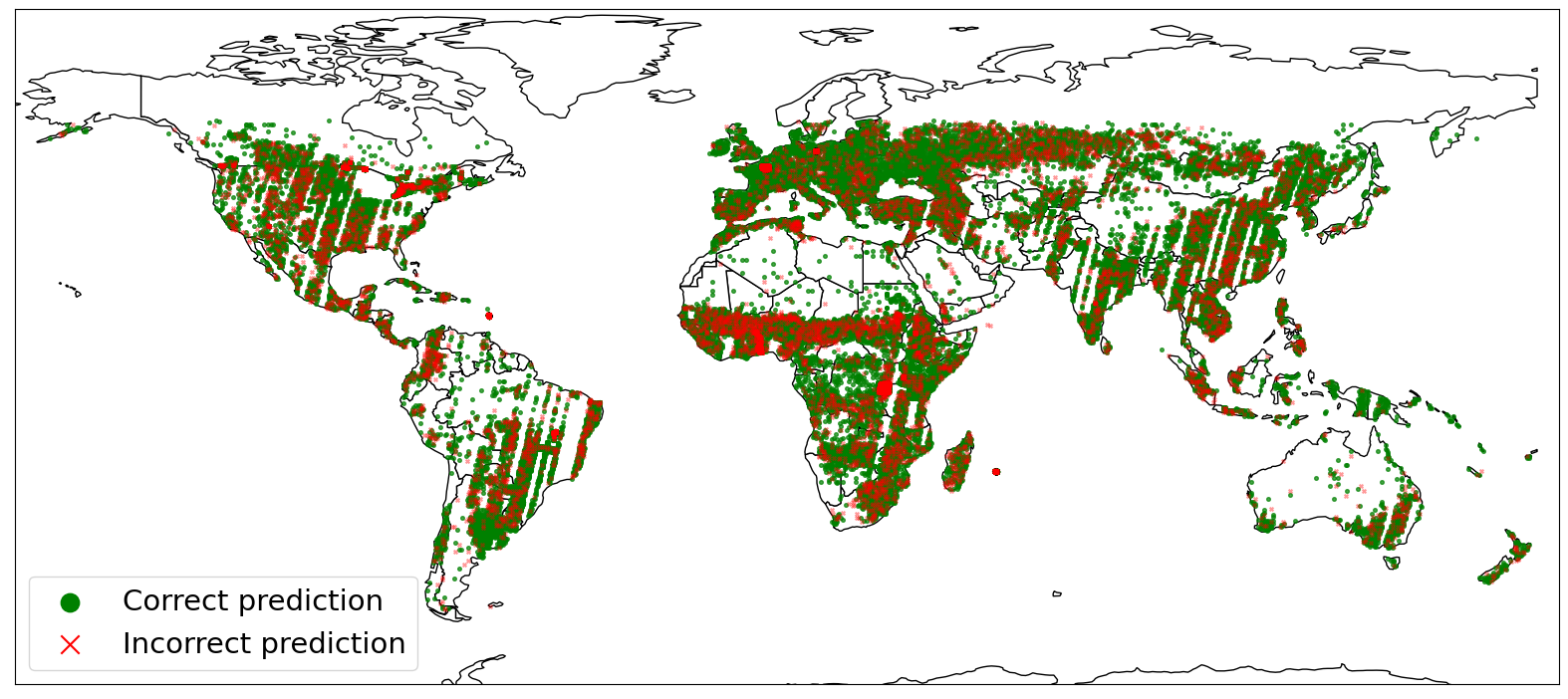}}
    \hfill
    \subfloat[MVL model with Hybrid gated strategy.]{\includegraphics[width=0.495\linewidth]{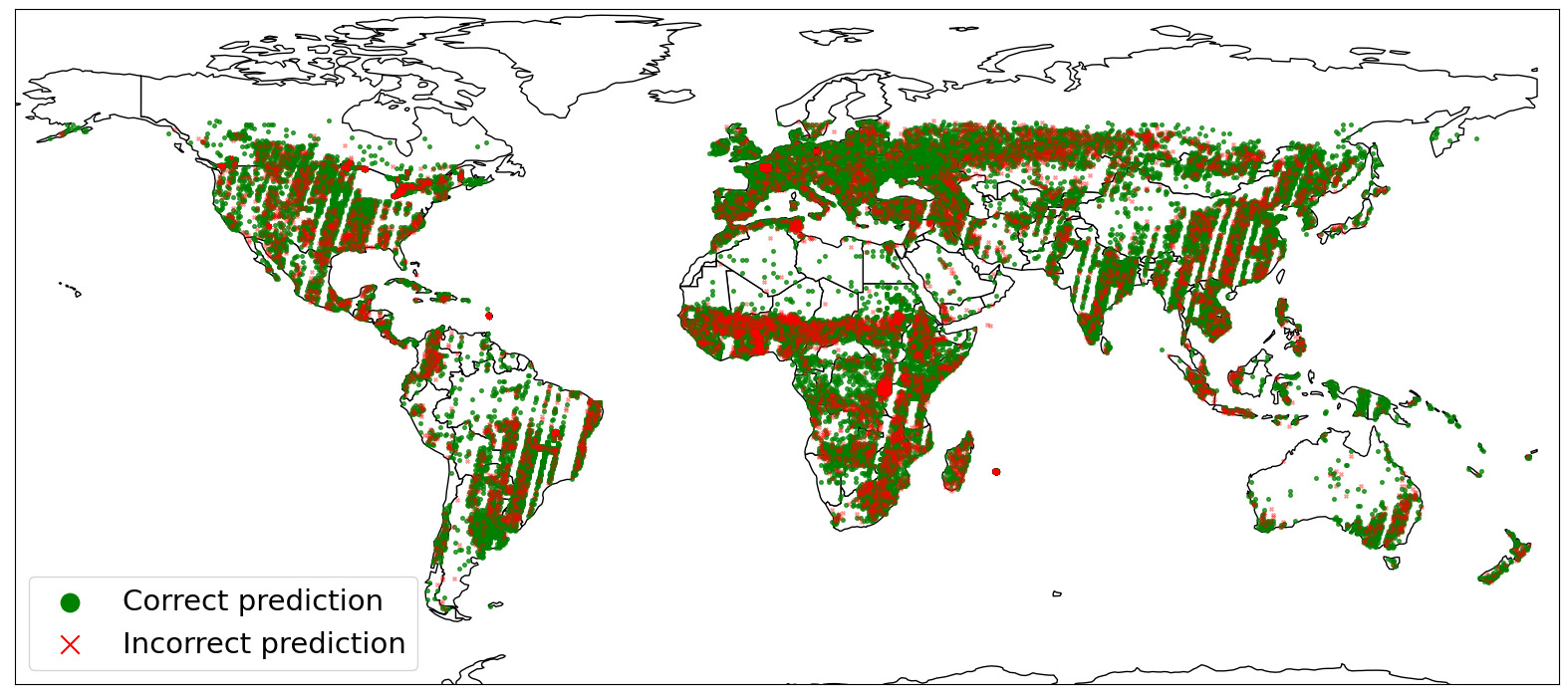}} \\
    \subfloat[MVL model with Ensemble strategy.]{\includegraphics[width=0.495\linewidth]{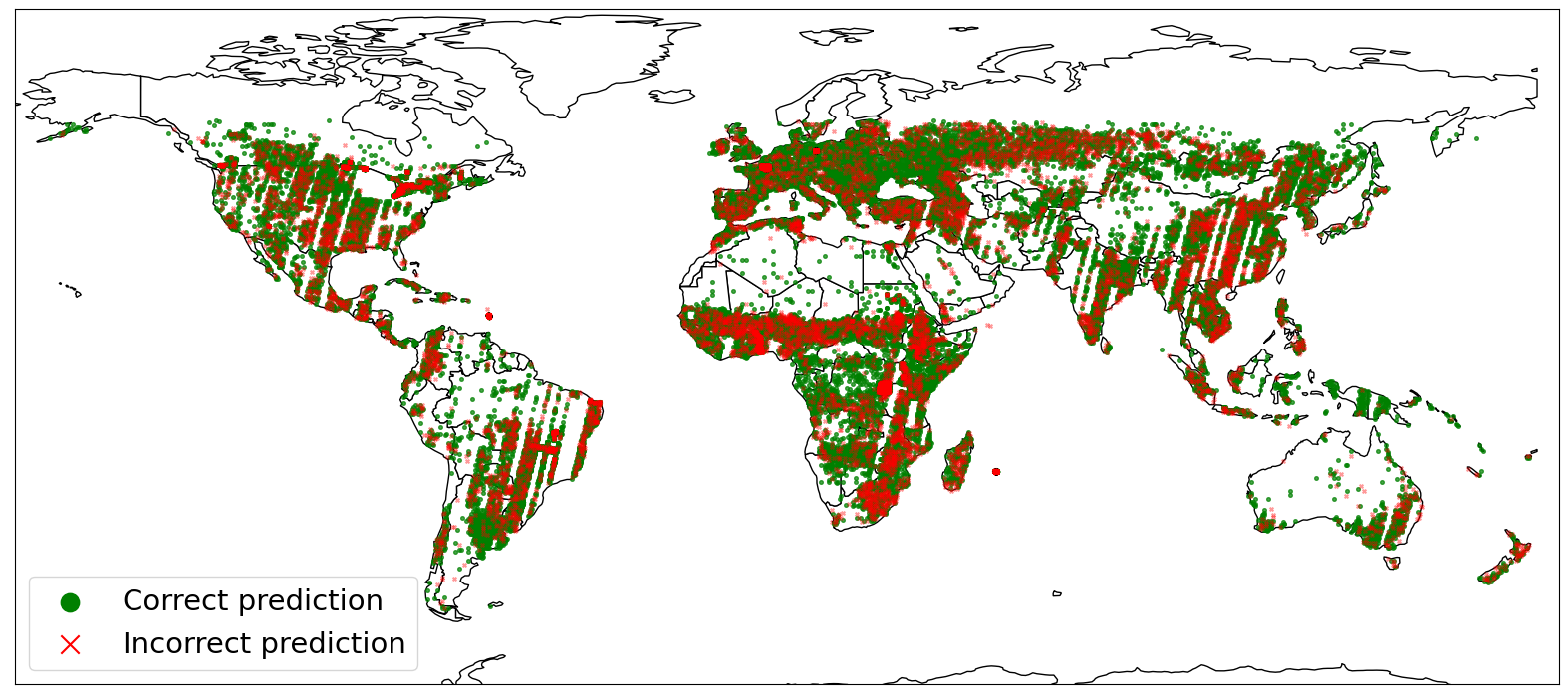}} \\  
    \caption{Predicted data points in the Global Binary evaluation for different fusion strategies using TempCNN as encoder. Red crosses are points with the predicted class equals to the target class, while green circles are samples with the predicted class different from the target.}\label{fig:exp_app:predictions}
\end{figure*}

\begin{figure*}[!t]
    \centering
    \subfloat[Single-view model fed with the optical view.]{\includegraphics[width=0.495\linewidth]{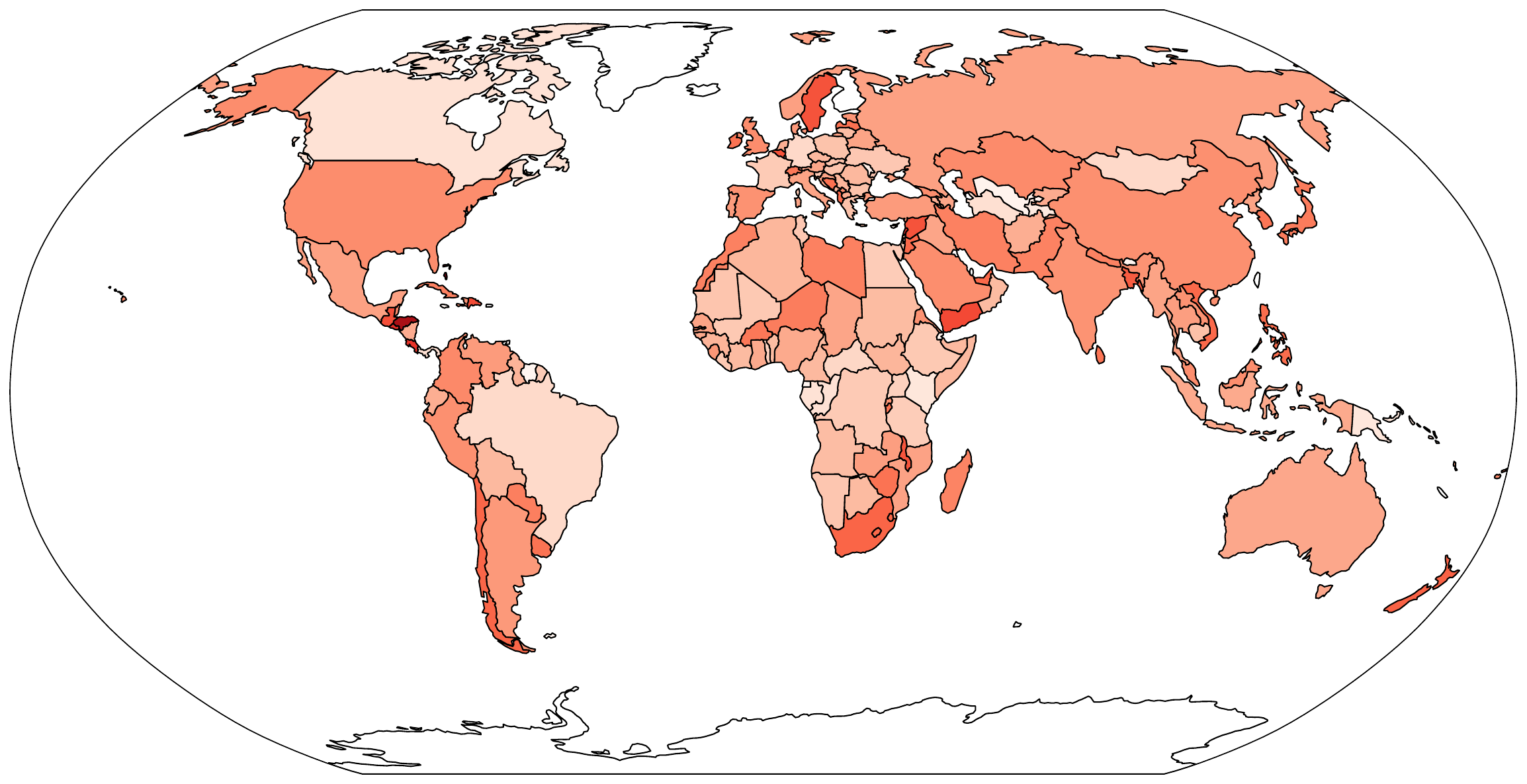}}
    \hfill
    \subfloat[MVL model with Input strategy.]{\includegraphics[width=0.495\linewidth]{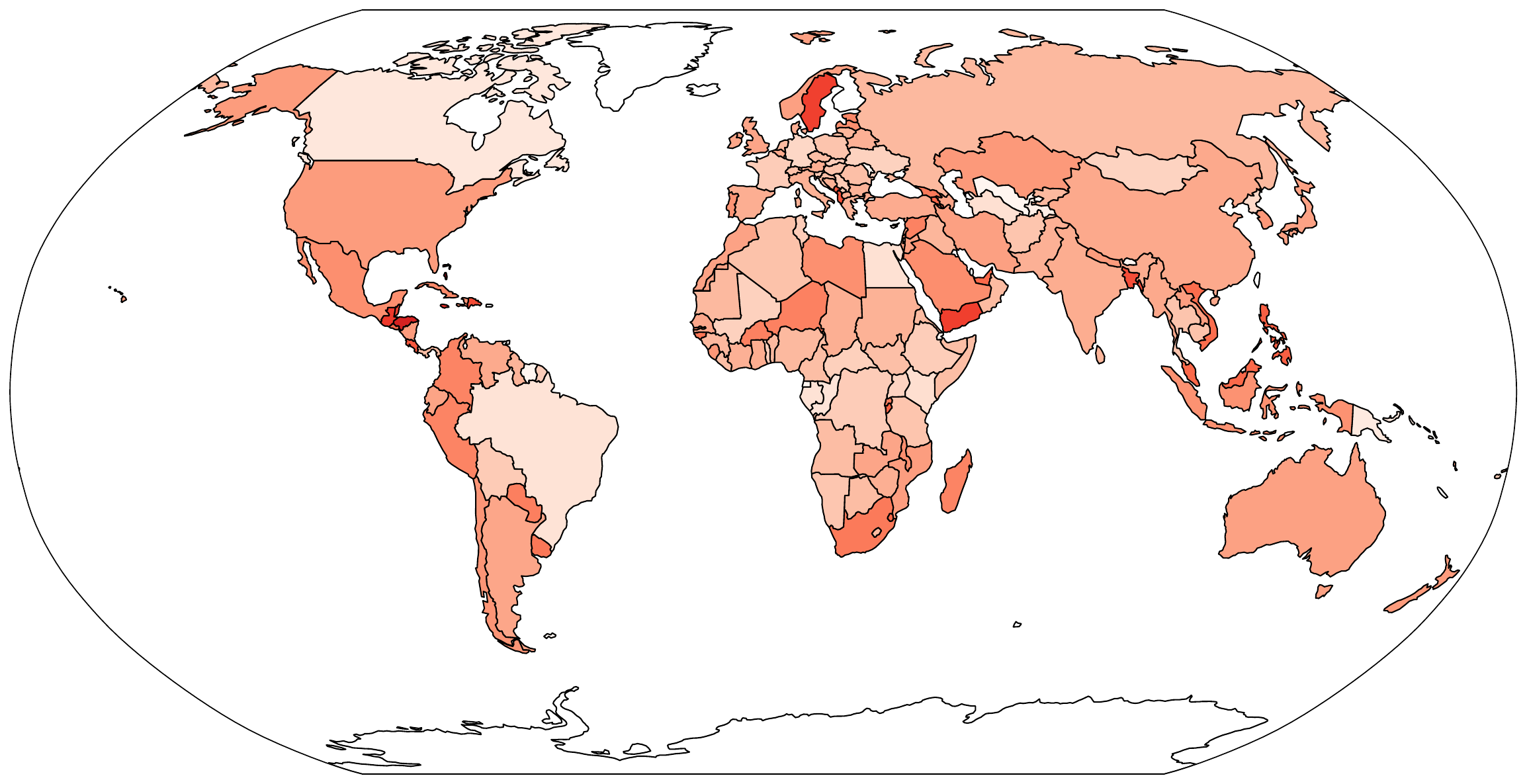}} \\\
    \subfloat[MVL model with Decision strategy.]{\includegraphics[width=0.495\linewidth]{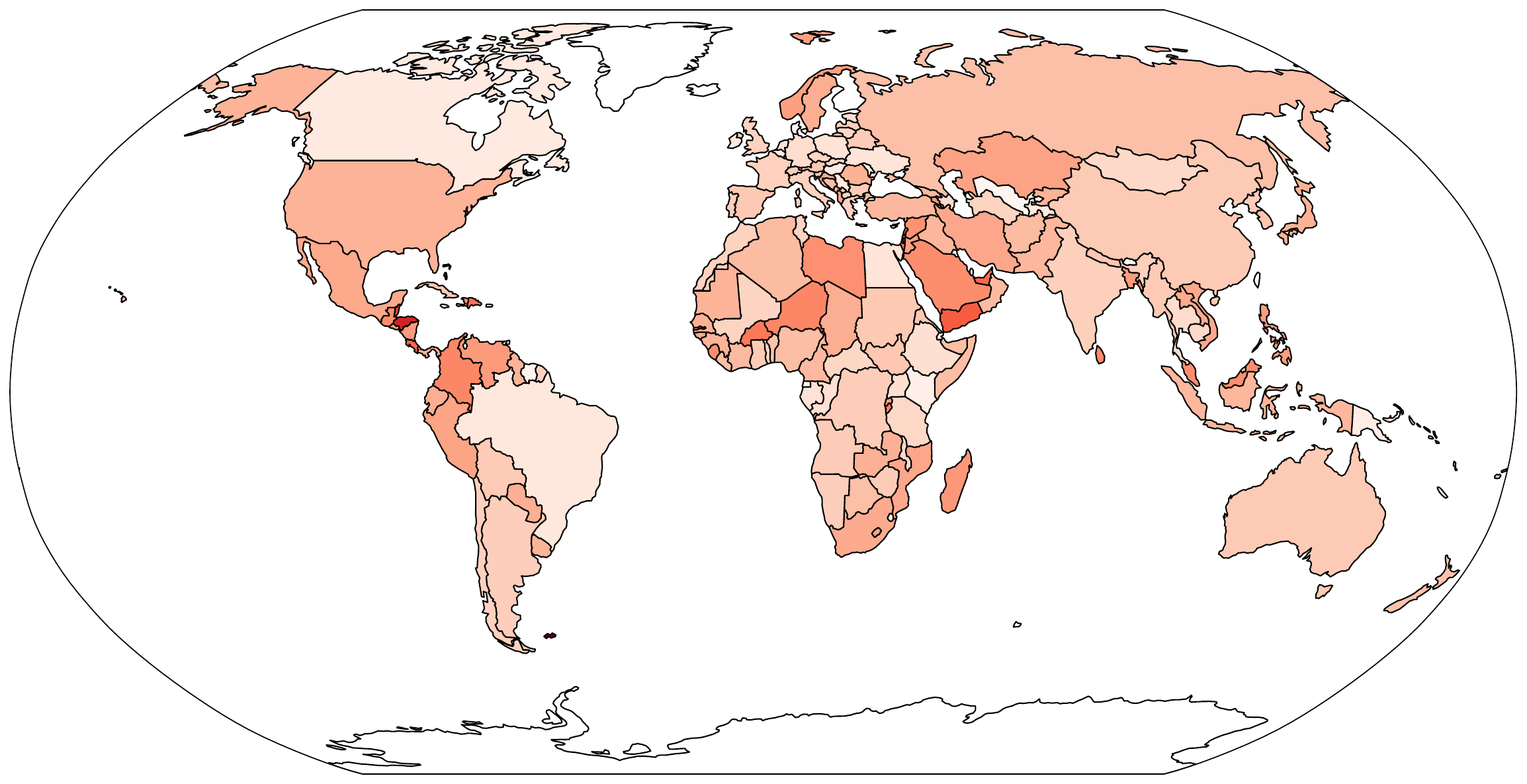}}
    \hfill
    \subfloat[MVL model with Hybrid gated strategy.]{\includegraphics[width=0.495\linewidth]{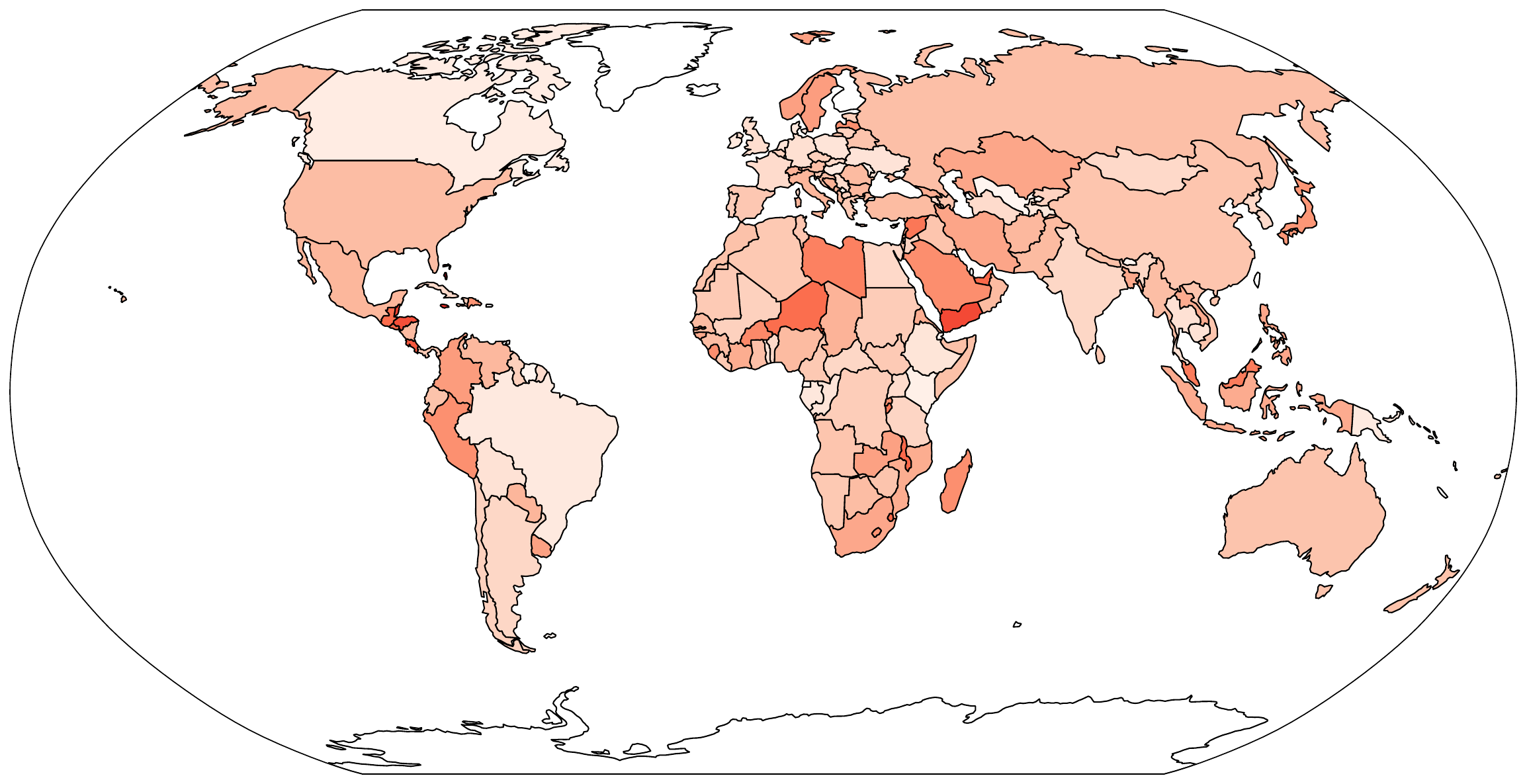}} \\
    \subfloat[MVL model with Ensemble strategy.]{\includegraphics[width=0.495\linewidth]{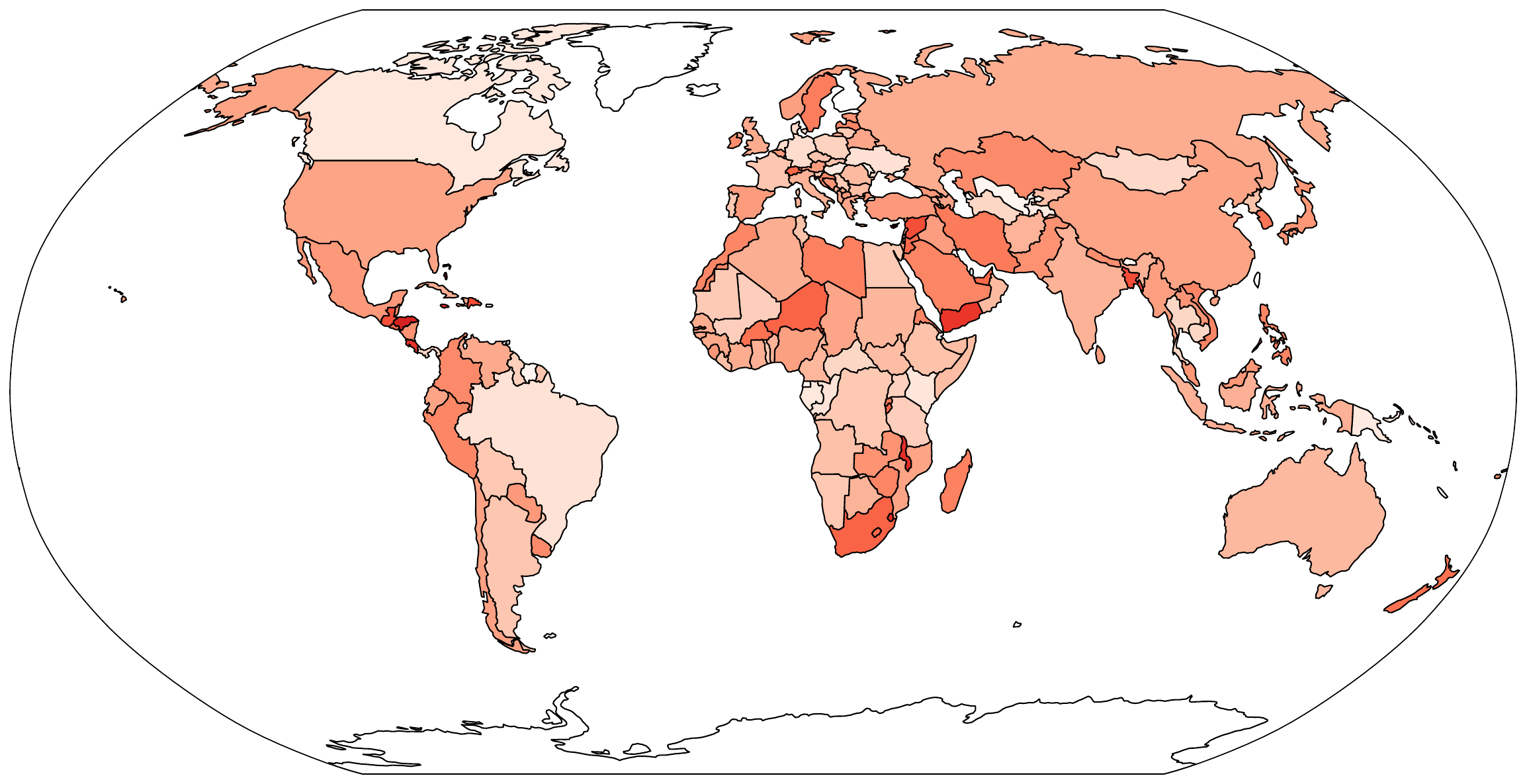}} \hfill
    \subfloat[MVL model with Feature gated strategy.]{\includegraphics[width=0.495\linewidth]{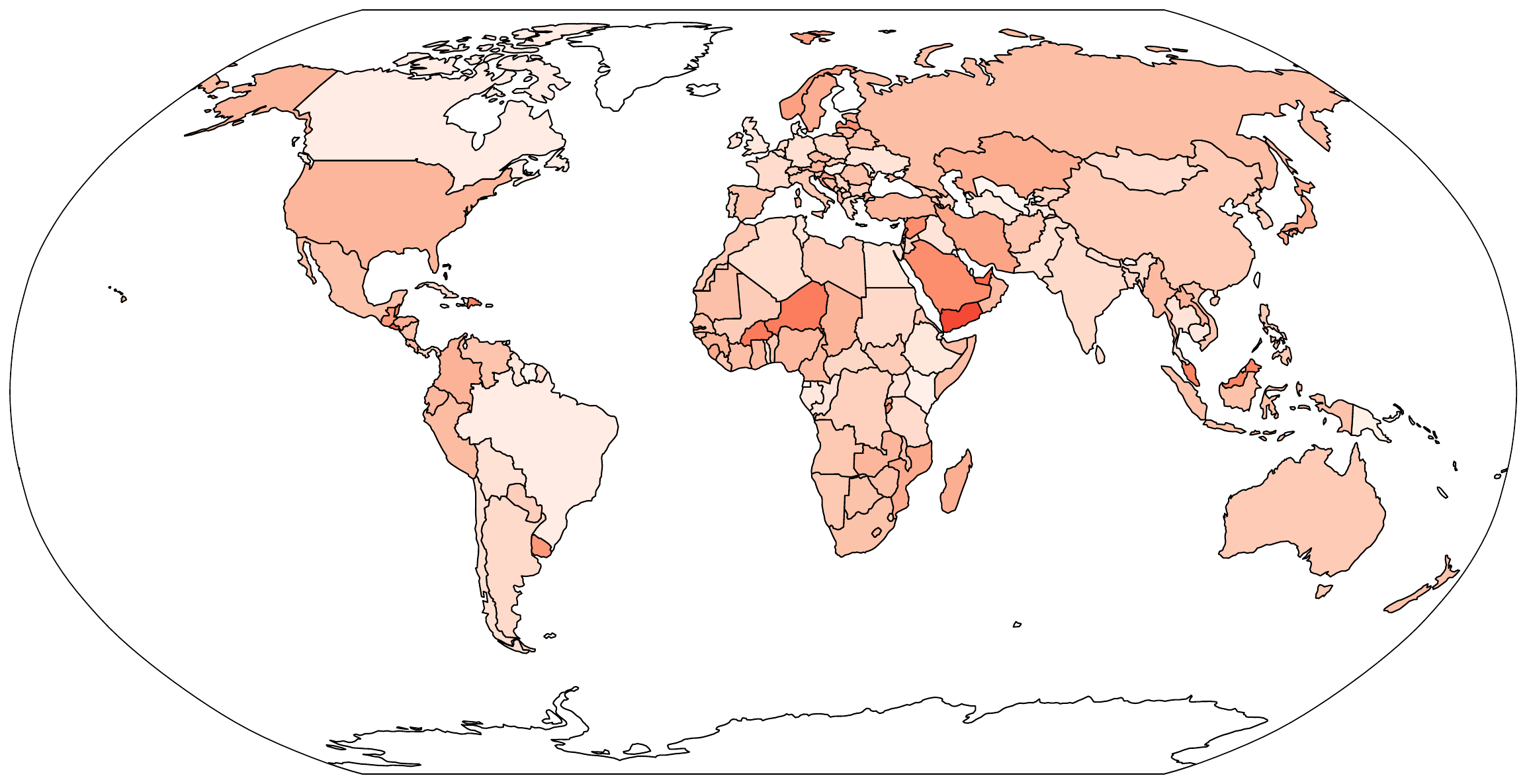}} 
    \caption{Error heatmap in the Global Binary evaluation for different fusion strategies using TempCNN as encoder. A stronger red color means more errors in the specific country.}\label{fig:exp_app:error_country}
\end{figure*}

We include additional figures to show the classification results in the global binary evaluation around the globe. For this, the predictions for the different fusion strategies using the TempCNN as encoder are shown in Fig.~\ref{fig:exp_app:predictions}. Additionally, an error heatmap per country is presented in Fig.~\ref{fig:exp_app:error_country} with all the fusion strategies using the TempCNN encoder architecture. 
For each MVL model, we observe different errors per region, which illustrates that the fusion strategies achieve different spatial advantages.

